\documentclass[11pt]{article}
\usepackage{amsfonts}
\usepackage{float}
\usepackage{pifont}
\usepackage{placeins} 
\usepackage[a4paper,margin=1in]{geometry}
\usepackage{booktabs}
\usepackage{array}
\usepackage{ragged2e}

\usepackage{hyperref}

\newcolumntype{L}[1]{>{\RaggedRight\arraybackslash}p{#1}}

\usepackage[final]{acl}
\usepackage{amsmath}
\usepackage{times}
\usepackage{latexsym}
\usepackage{booktabs}
\usepackage[T1]{fontenc}

\usepackage[utf8]{inputenc}

\usepackage{microtype}

\usepackage{inconsolata}

\usepackage{graphicx}

\usepackage{placeins}

\usepackage{subcaption}

\usepackage{placeins}

\usepackage{tabularx}
\usepackage{booktabs}
\usepackage{fvextra}
\usepackage{multirow}
\title{The Geometry of Metaphor:  Probing and Steering Metaphor Representations in Large Language Models}

\author{Shihui Li\thanks{Joint first authors}, Mette Andersen$^*$, Martha Lewis \\
        Institute for Logic, Language, and Computation \\ University of Amsterdam\\
        \texttt{shihui.li@student.uva.nl, \{m.f.andersen, m.a.f.lewis\}@uva.nl}}

\begin{document}
\maketitle
\begin{abstract}
Metaphor is a crucial aspect of human cognition. As such, we would like to understand how it is represented in large language models (LLMs) and how it might be controlled. We use mechanistic interpretability techniques to identify causal subspaces in a range of LLMs covering different architectures and sizes. We show that we are able to steer output on classification tasks across different parts of speech. We are also able to steer output on generation tasks, indicating that these techniques can be used to control the metaphoricity of LLM output.
\end{abstract}

\section{Introduction}
Metaphor is an essential part of everyday human language use. Sources such as \citet{steen2010metaphor, shutova-2010-automatic} show that metaphor pervades language. Furthermore, metaphor forms a basis for the fundamental creativity and concept formation that we are capable of \cite{lakoff2008metaphors}. Large language models such as Llama 3 \cite{grattafiori2024llama} have some capacity to understand and produce metaphorical language, although whether they are up to human performance is debated \cite{chakrabarty2021figurative, stowe-etal-2022-impli, ichien2024large}.

One way to understand the capacities of large language models is via \emph{mechanistic interpretability}: by examining the internals of large language models via, for example, the Logit Lens\footnote{\url{https://www.lesswrong.com/posts/AcKRB8wDpdaN6v6ru/interpreting-gpt-the-logit-lens}}, causal interventions \cite{meng2022locating}, or circuit discovery \cite{conmy2023towards}.  \citet{sharkey2025open} provides a thorough review and discussion of open problems in the area. One recent paper, \citet{marks2024geometry}, argued that Llama models exhibited a `truth direction' at particular layers, which was common to different datasets, and that the models' output could be steered by causal intervention on this layer.

Finding such a direction for metaphor would give insight into how models represent metaphor. This has been investigated to some extent \cite{boisson2024metaphors} but not yet at the mechanistic level. Finding this direction could also provide a useful `dial' that could potentially be used to steer the metaphoricity or literalness of output text, using techniques from e.g. \citet{li2023inference}. In this paper, we use activation patching \cite{meng2022locating} and distributed alignment search \cite{geiger2024finding} to find a `metaphoricity direction' or subspace within a given model. We show that these directions are causally effective: we are able to change the output of models in classification tasks for metaphor identification and interpretation; and in a generation task we are able to steer the model towards more metaphorical output.

\paragraph{Contributions}
\begin{itemize}
    \item We identify directions and subspaces for metaphoricity in a range of models, and show that these have some causal effect.
    \item We show that these directions are somewhat consistent across different metaphor datasets.
    \item We test seven models from four different model families (Gemma \cite{team2024gemma}, Qwen \cite{bai2023qwen}, Llama \cite{grattafiori2024llama}, Mistral \cite{jiang2023mistral7b}) across small (7-9B) and medium-large (27-70B) sizes.
    \item We show that steering models in a generation task can change model output from more literal to more metaphorical.
\end{itemize}

\section{Background and Related work}
\paragraph{Metaphor identification and interpretation}
In metaphor processing, research has focussed on two main tasks: \textit{metaphor identification}, that is, identifying whether a word is metaphorical, and \textit{metaphor interpretation}, that is, identifying the meaning of a metaphorical utterance. Work in using LLMs for metaphor identification include \citet{fuoli2026metaphor, reimann-scheffler-2025-using}. On the metaphor interpretation side, there has been a range of work showing that modern LLMs have relatively strong performance at understanding metaphor, especially when fine-tuned \cite{ichien2024large, liu-etal-2022-testing}.
Since metaphor is a key part of human language, we would like to understand how LLMs interpret and identify metaphor.

\paragraph{Activation Patching}
Mechanistic interpretability techniques attempt to understand and affect the output of large language models by inspecting their internal representations, for example by looking at the vectors of activations  produced at each layer of the model \cite{meng2022locating,marks2024geometry}. A common technique is to then \emph{causally intervene} on these representations. This means that we examine how a model's output changes when we alter its internal representations, for example by \emph{activation patching}, introduced in \citet{meng2022locating}; that is, we save the activations for one input, and patch these into the activations for another input, examining which layers and tokens affect the output.

In \citet{marks2024geometry}, the authors use activation patching to find a combination of layer and token input that represents a concept of truth in a language model.

\paragraph{Distributed Alignment Search (DAS)}\cite{geiger2024finding} is a method for finding causal concepts in model activations, based on the assumption that full-vector activations may be polysemantic, i.e. may be activated for different concepts. The method has been used in existing benchmarks \cite{mueller2025mib} for causal variable localization, and compared to other methods (SAEs, PCA, full-vector), DAS has been shown to find concepts that achieve superior causal performance. In this paper, we use DAS to find a causal concept of metaphoricity in the model's residual stream, and compare the performance of steering using DAS with the existing mass-mean steering used by \citet{marks2024geometry}.

\paragraph{Controlling Generation}
Controlling the style of LLM generated text is also an active area of research. Techniques used include training a separate model \cite{liu-etal-2022-semi}, or prompt-based style control such as in \cite{liu2024step}. Relatedly, \citet{li2023inference} alter the truthfulness of generated text via inference-time interventions.

\paragraph{Summary} Metaphor identification, interpretation, and generation are key aspects of language use. So far, in LLMs, these abilities have mostly been investigated by looking at behavioural outputs. In this paper, we use activation patching and distributed alignment search to investigate firstly whether metaphorical and literal sentences are linearly separable in activation space, secondly whether we can steer models towards literal or metaphoric output, and lastly whether the steering vectors are common to different datasets. We control the metaphoricity of generated output using the inference-time intervention technique introduced by \citet{li2023inference}. To our knowledge, this is the first attempt to control the metaphoricity of LLM output via intervention on the activations, rather than via prompting or building an external model.

\section{Datasets and Tasks}
\label{dataset area id}
\subsection{Datasets}
We use three datasets covering two different parts of speech (nouns and adjectives)\footnote{We initially also used a verb dataset, however model performance on this dataset was not sufficiently high to probe with interpretability techniques.}. Using mechanistic interpretability techniques requires that models are able to perform to a high standard on the task, and the structure of the dataset also needs to be fairly controlled. We therefore use relatively simply structured datasets. Following \citet{marks2024geometry}, we use datasets with the following properties: 1) each sentence contains exactly one metaphorical target word; 2) sentences are short and structurally simple; 3) within each dataset, sentences have the same syntactic structure.

We use the following three datasets. Examples of sentences with metaphorical nouns and adjectives are shown in Table~\ref{tab:dataset}.

\begin{table}[t]
\centering
\begin{tabular}{ll}
\hline
Category & Example sentence\\
\hline
Noun & The book is a \textcolor{blue}{puzzle}.\\
Adj  & The welcome is \textcolor{blue}{warm}.\\
\hline
\end{tabular}
\caption{Example sentences with metaphorically used nouns and adjectives. In example sentences, metaphorical target words are coloured blue.}
\label{tab:dataset}
\end{table}
\paragraph{Noun dataset (NOUN).}  
We start from the small training set of Fig-QA \cite{liu-etal-2022-testing}, which contains 200 human-written creative metaphors. Each example is paired with one correct rephrase and one incorrect rephrase in a Winograd-style format. Every two metaphor examples form a pair representing the meaning of a particular rephrase.

We first remove idioms, multi-clause descriptions, and long narrative sentences (e.g., \textit{``The painting has the creativity of Rembrandt and Picasso getting their paintings married''}). We then use ChatGPT \cite{openai2025chatgpt} to rewrite the remaining examples into short, direct metaphors with the parallel structure \textit{``The noun$_1$ is (a/an) noun$_2$''}. For example, the original sentence \textit{``The bear was the size of a whale after eating a smaller whale''} is simplified to \textit{``The bear is a whale''}. In each sentence, noun$_2$ metaphorically describes noun$_1$. We manually verify that each sentence is grammatically correct and preserves the original metaphorical mappings. The final dataset contains 130 sentence tuples. In the rest of the paper we refer to this dataset as \textbf{NOUN}.

\paragraph{Adjective dataset (ADJ).}  
We use the ADJ+N metaphor/literal dataset proposed by \citet{tsvetkov2014metaphor}. The test set contains 100 adjective–noun pairs with both literal and metaphorical usages. 
We modify this dataset to have a sentence format using the template \textit{``The N is ADJ.''} 
In this dataset, the adjective carries the metaphorical meaning and modifies the noun. In the remainder of this paper, we refer to this dataset as \textbf{ADJ}.

\paragraph{Domain-specific adjective dataset (DOM).}
We also use the dataset by \citet{gutierrez-etal-2016-literal} for DAS training, steering prediction and steering generation. The dataset consists of adjective-noun pairs across 8 different domains (clarity, depth, light, strength, substance, taste, temperature, texture). Each sample is labelled literal (e.g. ``warm coffee'') or metaphorical (e.g. ``warm welcome''). We evaluate on n=100 per steering direction from the held-out set not used for training DAS. In the remainder of this paper we refer to this dataset as \textbf{DOM}.

\subsection{Tasks}
\label{sec:tasks}
We carry out an \textbf{identification} task, an \textbf{interpretation} task, and a \textbf{generation} task.

\paragraph{Identification}
Models are prompted with a four-shot setup to label a sentence as either `literal' or `metaphoric'. The adjective datasets from \citet{tsvetkov2014metaphor} and \citet{gutierrez-etal-2016-literal} are in the form \texttt{adjective noun}. We rephrase these into sentences of the form `The \texttt{noun} is \texttt{adjective}'. The full prompt for \textbf{ADJ} is then as follows. See Appendix \ref{sec:noun_prompt} for the prompt for \textbf{NOUN}
\\
\
\\
    The town is sleepy. This sentence is: metaphoric \\
    The sand is white. This sentence is: literal \\The moon is young. This sentence is: metaphoric \\
    The banana is ripe. This sentence is: literal 
    \\\texttt{sentence}. This sentence is:

\paragraph{Interpretation}
Models are prompted with a four-shot setup. In each shot, a metaphorical sentence and a candidate paraphrase are given, and the paraphrase is labelled as `apt' or `inapt'. This task uses \textbf{NOUN}, which is based on FigQA \cite{liu-etal-2022-testing}. Apt/inapt pairs are taken from their setup. The full prompt is given below:
\\
\
\\
Determine whether the paraphrase is apt or inapt.\\
The politician is glass: The politician is transparent. This paraphrase is: apt.\\ 
The politician is a wall: The politician is transparent. This paraphrase is: inapt.\\
The cake is heaven: The cake is not delicious. This paraphrase is: inapt.\\
The cake is heaven: The cake is delicious. This paraphrase is: apt.\\
\texttt{sentence}: \texttt{rephrase}. This paraphrase is:

\paragraph{Generation}
Models are prompted with a five-shot setup to output an interpretation of a sentence that could be either literal or metaphorical. The dataset used is \textbf{DOM}, and we reword adjective-noun pairs into sentences of the form `The \texttt{noun} was \texttt{adjective}'. The full prompt is given below:
\\
\
\\
The welcome was warm. The meaning of `warm' is: affectionate and emotionally inviting.\\
The lamp was bright. The meaning of `bright' is: emitting a strong, clear light.\\
The novel was deep. The meaning of `deep' is: intellectually complex and rich in meaning.\\
The stone was smooth. The meaning of `smooth' is: having an even surface with no roughness.\\
The defeat was bitter. The meaning of `bitter' is: painful, hard to accept, and leaving resentment.\\
The \texttt{noun} was \texttt{adj}. The meaning of `\texttt{adj}' is:

\section{Methods}
\subsection{Identifying areas encoding metaphoricity}

For each of the datasets and tasks described above, we use one of two techniques designed to isolate the part of the model used to encode metaphoricity. For the identification and interpretation tasks across \textbf{NOUN}, \textbf{ADJ}, and \textbf{DOM}, we use \textbf{activation patching}. For \textbf{DOM} and for the generation task, we also use \textbf{distributed alignment search (DAS)} to identify a linear subspace encoding metaphoricity. We use the \textbf{mass-mean probe} from \citet{marks2024geometry} and the learned DAS subspaces to steer model output in the identification and interpretation tasks, requiring single-token output. Finally, we use the mass-mean probe and DAS subspaces to steer generation over multiple tokens, finding that it is possible to steer models towards more metaphorical or more literal output.

\subsubsection{Activation Patching}
In the identification task, we use the following procedure. On \textbf{ADJ}, we insert the following sentence into the prompt template in Section \ref{sec:tasks} to form a prompt $p_L$= \textit{The skin is dry.}
For prompt $p_L$ we expect the output `literal'. We then form a paired prompt $p_M$ replacing \textit{skin} with \textit{wit}. For this prompt we expect the output `metaphoric'. To find out which hidden states contain information about the metaphoricity of the sentence, we run the model on the input $p_M$ and cache the activations $h_{i,\ell}(p_M)$ for each token position $i$ and layer $\ell$. Then, for each $i$ and $\ell$, we run the model on $P_L$ and modify the model's forward pass by swapping $h_{i,\ell}(P_L)$ for $h_{i,\ell}(P_M)$. Finally, we record the difference in log probability between the tokens "metaph"\footnote{LlaMA-3.1-8B tokenizes "metaphor" into subwords "metaph" and "or". In our study, we examine the model's first output token "metaph".} and "literal". The larger the difference, the more influential the hidden state in position $i$ and layer $\ell$ on the model's prediction. For the dataset \textbf{NOUN}, we use the prompt pair $p_L$=\textit{The breeze is cold} and $p_M$=\textit{The breeze is ice}.
For the interpretation task, we use the prompt pair $p_{inapt}$ = \textit{The puzzle is rocket science: The puzzle is easy} and $p_{apt}$ = \textit{The puzzle is rocket science: The puzzle is difficult}. 

\subsubsection{Distributed Alignment Search}
We use Distributed Alignment Search (DAS) \cite{geiger2024finding} to localize, if it exists, a low-dimensional, linear subspace encoding metaphoricity. 
After identifying the most causally implicated token-position in the residual stream using full-vector activation patching, DAS trains a rotation matrix on top of these activations. The rotation matrix is trained to maximize Interchange Intervention Accuracy (IIA), which is the fraction of samples where patching flips model prediction in the expected direction.

Concretely, let $\mathbf{h} \in \mathbb{R}^d$ denote the residual stream at a fixed layer $\ell$ and token position $p$. As will be seen in Figure \ref{fig:patching-task1}, the full stop token at the end of the target sentence in each prompt is highly causally implicated. 
We perform DAS at this token at the layer that optimises the IIA score.

DAS learns a matrix $\mathbf{R} \in \mathbb{R}^{d \times k}$ with orthonormal
  columns ($\mathbf{R^\top R} = \mathbb{I}_k$), defining a $k$-dimensional subspace of the representation space. Given a source hidden
  state $\mathbf{h}^{\text{source}}$ and a base hidden state $\mathbf{h}^{\text{base}}$, the intervention replaces the
  subspace component of $\mathbf{h}^{\text{base}}$ with that of $\mathbf{h}^{\text{source}}$:

 \begin{equation*}
\mathbf{h}_{\text{patched}} = \underbrace{\mathbf{h}}_{\substack{\text{original}\\\text{activation}}} - \underbrace{\mathbf{RR^\top h}}_{\substack{\text{remove original}\\\text{subsp. coords}}} + \underbrace{\mathbf{RR^\top} \boldsymbol{\mu}_{\text{target}}}_{\substack{\text{inject target}\\\text{subsp. coords}}}
\end{equation*}

During training, we alternate intervention directions: on even steps we patch literal representations into metaphor prompts
  (aiming to flip the output to literal), and on odd steps the reverse. The loss for a single example is $  \mathcal{L} = \text{logit}(\text{other}) - \text{logit}(\text{target})$
where target is the correct post-intervention label and other is the original label. $\mathbf{R}$ is updated via Adam for 1,000 steps while the model weights are frozen.

IIA is the fraction of samples that are correctly flipped. In our case, if $\mathbf{h}^{\text{base}}$ is predicted as ``metaphor'' for the model, it should flip to ``literal'' post-patching to contribute positively to IIA. Hence, IIA = 1.0 means we have found the variable capable of flipping \textit{all} held-out examples.

\subsection{Model steering}
We evaluate whether we can steer incorrect predictions towards correct ones using two methods: full-vector mass-mean steering, as used by \citet{marks2024geometry}, and steering using the mean activations for metaphor-sample and literal-sample activations in the DAS subspace. The means are computed from correctly predicted samples, and we steer incorrectly predicted samples in both directions.

Mass-mean steering is implemented as:
  \begin{equation}
  \label{eq:mm}
    \mathbf{h}_{\text{steered}} \leftarrow \mathbf{h} + \alpha(\boldsymbol{\mu}_{\text{met}} - \boldsymbol{\mu}_{\text{lit}})
\end{equation}

DAS steering is implemented as:.
\begin{equation}
    \label{eq:das}
    \mathbf{h}_{\text{steered}} \leftarrow \mathbf{h} - \alpha\,\mathbf{R}\mathbf{R}^\top \mathbf{h} + \alpha\,\mathbf{R}\mathbf{R}^\top \boldsymbol{\mu}_{\text{target}}
\end{equation}

We set $\alpha=1$ for the identification and interpretation tasks on \textbf{NOUN}, \textbf{ADJ}, and \textbf{DOM}. 
On \textbf{DOM} we train DAS matrices firstly across the whole dataset, and subsequently for each individual domain.

\subsubsection{Generation of multiple tokens}
To test whether the DAS subspace encodes a general-purpose concept of metaphor, which extends beyond the prediction task it was trained on, we steer free-form generation. We prompt the model as described in section \ref{sec:tasks}. During decoding, we steer every generated token as in \citet{li2023inference}: we use a hook function to save the activation we use for patching, and inject it at every generated token.
 
We compare baseline generation, mass mean, and DAS outputs across $\alpha \in \{1, 2, 5, 10\}$ and both steering directions (lit $\to$ met and met $\to$ lit).

\paragraph{Evaluation using LLM-as-judge}

To quantify the degree to which generated output changed from explaining a literal use of an adjective to a metaphorical use and vice versa, we use Qwen 2.5 72B Instruct as a judge. Only the first sentence of each completion is judged, and a degeneracy filter applied before judging: a completion is marked degenerate if any single word repeats $\geq$ 3 times. The judge is prompted using the prompt shown in Appendix \ref{sec:judge-prompt}.

\begin{table*}[!htbp]
\centering
\resizebox{\textwidth}{!}{
\begin{tabular}{llcccccccc}
\toprule
\textbf{Task} & \textbf{Dataset} 
& \textbf{LLaMA-3.1-8B} 
& \textbf{LLaMA-3.1-70B} 
& \textbf{Qwen-2.5-7B} 
& \textbf{Qwen-2.5-72B} 
& \textbf{Mistral-7B} 
& \textbf{Mixtral-8x7B} 
& \textbf{Gemma-2-9B} 
& \textbf{Gemma-2-27B} \\
\midrule
\multirow{3}{*}{Identification}
 & \textbf{ADJ}  & 0.755 & 0.915 & 0.860 & 0.930 & 0.760 & 0.870 & 0.930 & 0.915 \\
 & \textbf{NOUN} & 0.840 & 0.905 & 0.815 & 0.888 & 0.645 & 0.881 & 0.873 & 0.900 \\
 & \textbf{DOM}  & 0.841 & 0.907 & 0.805 & 0.911 & 0.814 & 0.825 & 0.881 & 0.464 \\
\midrule
\multirow{1}{*}{Interpretation}
 & \textbf{NOUN} & 0.727 & 0.862 & 0.696 & 0.815 & 0.765 & 0.796 & 0.845 & 0.858 \\
\bottomrule
\end{tabular}
}
\caption{Baseline accuracy on the 4-shot evaluation on identification and interpretation tasks across all models and datasets. DOM results use the M\&T 5-shot prompt. Gemma-2-27B excluded from \textbf{DOM} analysis.}
\label{tab:tasks-acc}
\end{table*}

\paragraph{Metrics}

We report the following metrics: \textbf{Flip rate (MM / DAS)}: fraction of of completions where judge label matches the target direction. Degenerate responses are counted as 0, and we only consider the samples where the judge was able to judge correctly on the baseline generation pre-steering.

\subsection{Models Evaluated}
We evaluate seven instruction-tuned models across four model families\footnote{Where available, we also evaluated non-instruction tuned models, but found that their performance on the tasks was not strong enough for the application of interpretability techniques.} on metaphor identification and interpretation tasks. These are: Gemma 2 9B IT; LLaMA 3.1 8B IT; LLaMA 3.1 70B IT; Mistral 7B IT; Mixtral 8$\times$7B IT; Qwen 2.5 7B IT; Qwen 2.5 72B IT, where IT stands for intstruction tuned.

\section{Results}
\subsection{Baseline Performance} We evaluate all models on the identification and interpretation tasks across all datasets (see Table ~\ref{tab:tasks-acc}). We can see that all models are able to perform these tasks to a reasonable standard with the exception of Gemma 27B on \textbf{DOM}. This modes is excluded from further analysis on this dataset.

\begin{figure*}[t]
\centering
\begin{subfigure}{0.45\textwidth}
\includegraphics[width=\linewidth]{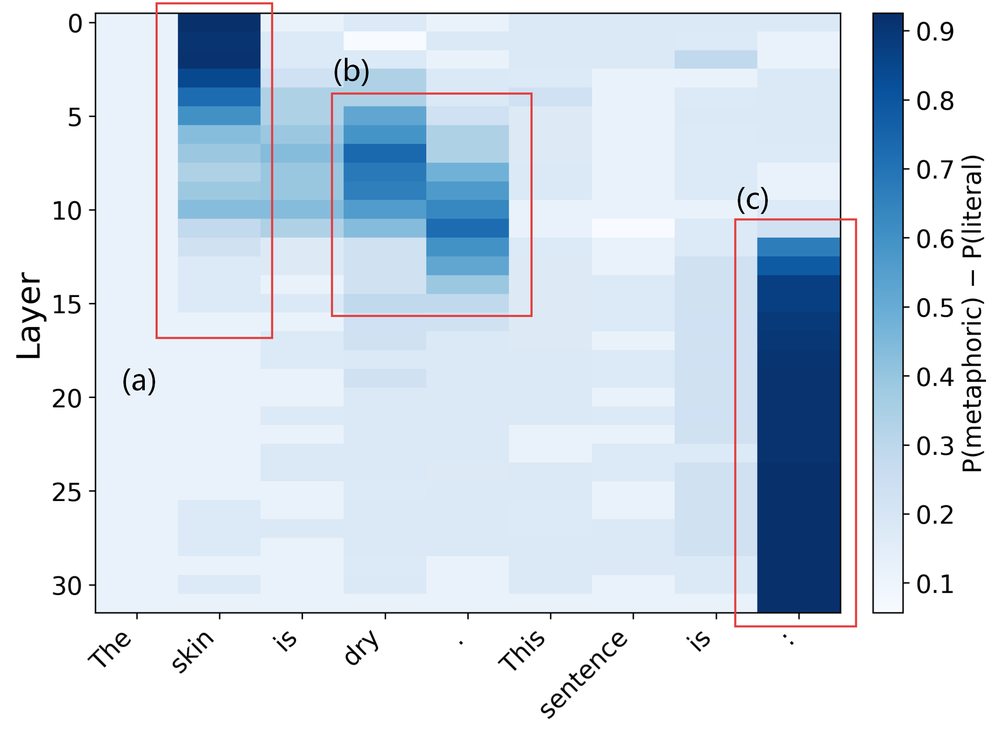}
\caption{Llama 8b \textit{identification} task}
\end{subfigure}
\hfill
\begin{subfigure}{0.45\textwidth}
\includegraphics[width=\linewidth]{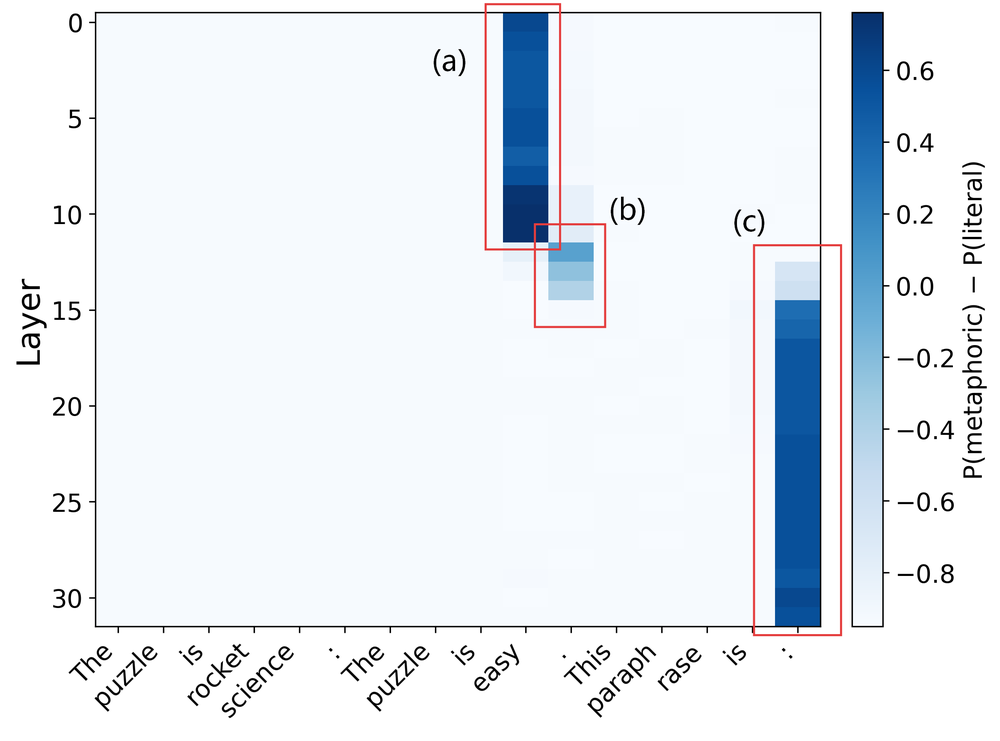}
\caption{Llama 8b \textit{interpretation} task}
\end{subfigure}
\hfill
\caption{\textbf{Activation patching results for Llama 3.1 8B.}
\textbf{(a) Identification task.}
The prompts use the literal sentence \textit{“The skin is dry.”} and the metaphorical sentence \textit{“The wit is dry.”} from \textbf{ADJ} We patch activations from the literal prompt into the metaphorical prompt. 
\textbf{(b) Interpretation task.}
The prompts use the metaphorical sentence \textit{“The puzzle is rocket science.”}, the correct paraphrase \textit{“The puzzle is easy.”}, and the incorrect paraphrase \textit{“The puzzle is difficult.”} from \textbf{NOUN}. We patch activations from the apt paraphrase into the inapt paraphrase.}
\label{patching exp}
\end{figure*}

\subsection{Identifying Metaphoricity}
\label{sec:iden}
\subsubsection{Activation patching}
Examples of patching from the interpretation and identification tasks are presented in Figure~\ref{patching exp}. Additional patching results are shown in Figures~\ref{fig:patching-task1} and~\ref{fig:patching-task2} in the Appendix.

Across both parts of speech and both tasks, we observe three groups of activations that most strongly influence the output (Figure~\ref{patching exp}). Two of these groups are expected. Group (a) corresponds to tokens that differ between the patching source and base prompts (e.g., “skin” and “wit”). Group (c) corresponds to the final token “:”, which is natural because the hidden states of the final token directly determine the next-token prediction.

Group (b) contains the final token “.” in the sentence or paraphrase, as well as the token immediately preceding it.  We hypothesize that Group (b) activations encode information about the metaphoricity of the sentence or paraphrase. 

\begin{figure*}
    \centering
    \begin{subfigure}{0.32\textwidth}
    \includegraphics[width=\textwidth]{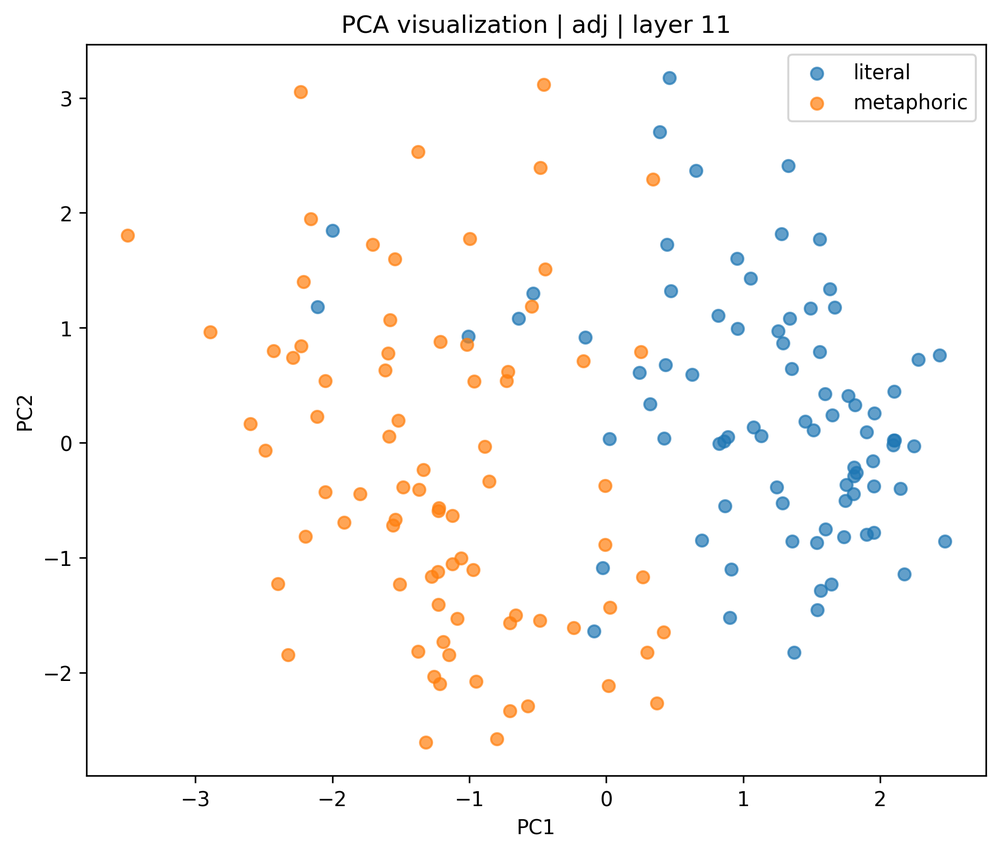}
    \end{subfigure}
    \hfill
    \begin{subfigure}{0.32\textwidth}
        \includegraphics[width=\linewidth]
{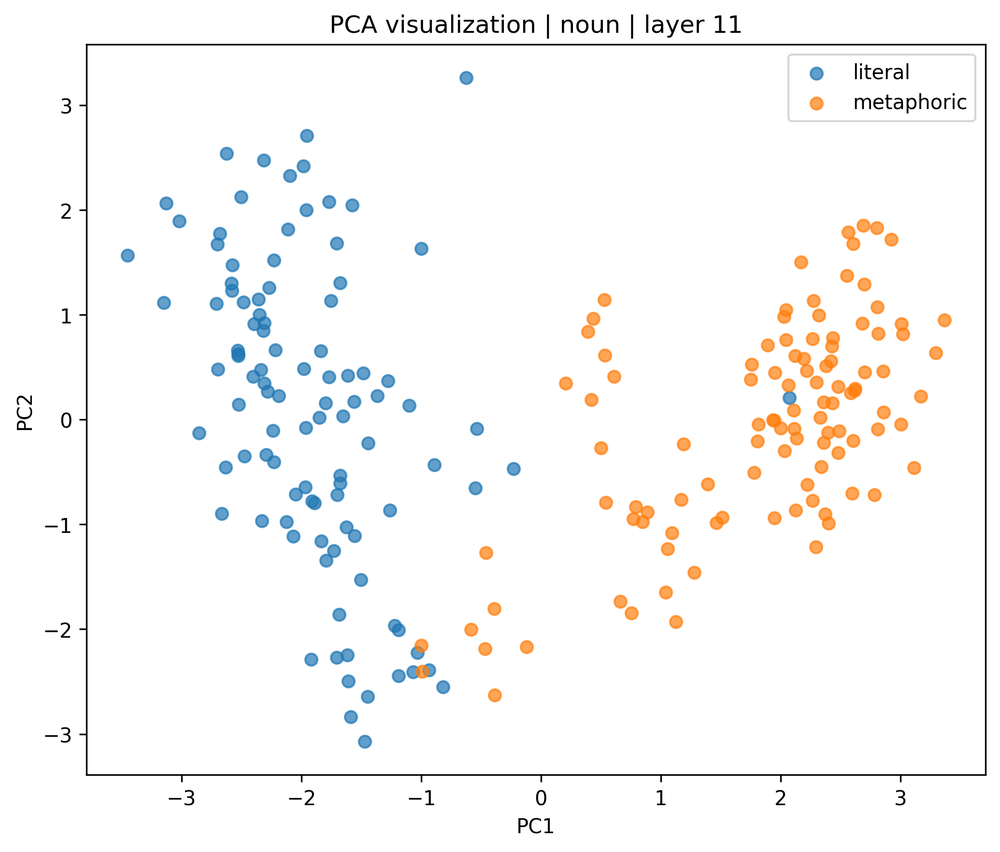}
    \end{subfigure}
    \hfill
    \begin{subfigure}{0.32\textwidth}
        \includegraphics[width=\linewidth]{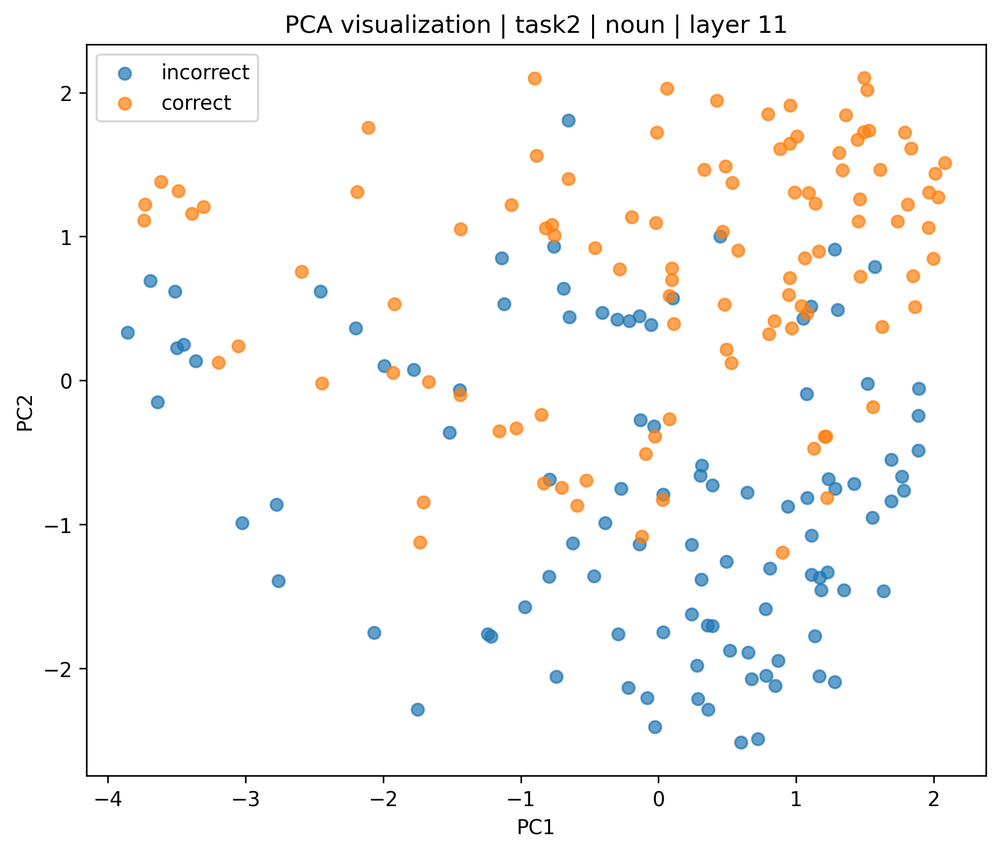}
    \end{subfigure}
    \caption{PCA visualization of activations at layer 11 for Llama 3.1 8B, for the identification task on the adjective (left) and noun(middle) datasets, and for the interpretation task on the adjective dataset (right). Notice that literal and metaphoric sentences are more or less linearly separable.}
    \label{fig:pcaplots_llama}
\end{figure*}

Based on the patching results, we extract the hidden states of the selected period token on the layer which has the most influence on the model's output (e.g.\ 11th layer for Llama-3.1-8B, 29th layer for Llama-3.1-70B). We visualize LLM representations of our datasets using principal component analysis (PCA). As shown in Figure \ref{fig:pcaplots_llama}, the first two principal components (PC1 and PC2) linearly separate metaphorical from  literal and apt from inapt across \textbf{NOUN} and \textbf{ADJ} and both tasks. More results are available in Figures~\ref{fig:pca-adj},\ref{fig:pca-noun} and \ref{fig:pca-task2-noun} in Appendix \ref{sec:PCAplots_allmodels}.

\subsubsection{DAS training}
\label{sec:DAS_training}
\begin{table}[htbp]
\centering
\caption{Global (all domains pooled) DAS IIA across models. IIA reported as mean $\pm$ std with 95\% CI.}
\label{tab:global-das-iia}
\small
\resizebox{\columnwidth}{!}{%
\begin{tabular}{lcccc}
\toprule
Model & $\ell$ & $k$ & {IIA $\pm$ std} & 95\% CI \\
\midrule
Gemma-9B      & 19 & 16 & $0.807 \pm 0.007$ & $[0.799,\ 0.815]$ \\
LLaMA-8B      & 11 & 16 & $0.793 \pm 0.006$ & $[0.783,\ 0.802]$ \\
Mistral-7B    & 10 & 32 & $0.711 \pm 0.019$ & $[0.698,\ 0.722]$ \\
Mixtral-8x7B  & 11 & 16 & $0.691 \pm 0.026$ & $[0.680,\ 0.701]$ \\
Qwen-7B       & 14 & 16 & $0.830 \pm 0.000$ & $[0.821,\ 0.839]$ \\
LLaMA-70B     & 28 & 32 & $0.640 \pm 0.025$ & $[0.630,\ 0.650]$ \\
Qwen-72B      & 52 & 32 & $0.843 \pm 0.024$ & $[0.836,\ 0.850]$ \\
\bottomrule
\end{tabular}
}
\end{table}

We select the optimal layer for training DAS using plain activation patching at the full stop token across layers. We extract the full hidden vector at this token at the optimal layer, and train two DAS matrices (\textit{k}=16 and \textit{k}=32). In Table \ref{tab:global-das-iia} we report best layer $\ell$, subspace dimension $k$ and IIA score.

We are interested in whether the model has a domain-general causal encoding of metaphor, or multiple domain-specific ones. We therefore compare the IIA for a global DAS matrix and multiple domain-specific DAS matrices on a held out evaluation set for each domain (not used for training global DAS, nor the domain-specific variant). Results for Llama 3.1 8B are reported in Table \ref{tab:das_domain_llama}. We see that the global DAS matrices have higher IIA than domain-specific matrices. For all results, see Table \ref{tab:domain-das-iia} and Table \ref{tab:transfer-iia}.

\begin{table}[t]
\centering
\small
\caption{LLaMA-8B: Domain-specific DAS IIA vs. global DAS transfer on held-out evaluation set}
\label{tab:das_domain_llama}
\setlength{\tabcolsep}{4pt}
\resizebox{0.45\textwidth}{!}{

\begin{tabular}{lcc}
  \hline
  Concept & Domain-specific DAS & Global DAS (transfer) \\
  \hline
  Clarity     & $0.316 \pm 0.070$ $[0.262, 0.369]$ & $0.500$ $[0.404, 0.596]$ \\
  Depth       & $0.343 \pm 0.010$ $[0.299, 0.386]$ & $0.910$ $[0.868, 0.952]$ \\
  Light       & $0.322 \pm 0.122$ $[0.294, 0.352]$ & $0.727$ $[0.674, 0.776]$ \\
  Streng.    & $0.606 \pm 0.061$ $[0.582, 0.628]$ & $0.967$ $[0.953, 0.982]$ \\
  Sub.   & $0.309 \pm 0.059$ $[0.248, 0.371]$ & $0.814$ $[0.714, 0.900]$ \\
  Taste       & $0.645 \pm 0.054$ $[0.588, 0.706]$ & $0.842$ $[0.763, 0.921]$ \\
  Temp. & $0.470 \pm 0.051$ $[0.435, 0.506]$ & $0.568$ $[0.500, 0.627]$ \\
  Tex.    & $0.343 \pm 0.022$ $[0.311, 0.376]$ & $0.713$ $[0.657, 0.765]$ \\
  \hline
  \end{tabular}
  }
\end{table}

\begin{table}[t]
\centering
\scriptsize
\setlength{\tabcolsep}{4pt}
\renewcommand{\arraystretch}{1.12}

\begin{tabular}{lllcc}
\toprule
Task & train/test & probe 
& \shortstack{\scriptsize met$\to$lit \\ \scriptsize /apt$\to$inapt}
& \shortstack{\scriptsize lit$\to$met \\ \scriptsize /inapt$\to$apt} \\
\midrule

\multirow{8}{*}{Task 1}
& \multirow{2}{*}{Adj/Adj}
& LogReg  & .038 & .051 \\
& & MM-diff & \textbf{.227} & \textbf{.471} \\
\cmidrule(lr){2-5}

& \multirow{2}{*}{Noun/Noun}
& LogReg  & .025 & .041 \\
& & MM-diff & \textbf{.195} & \textbf{.449} \\
\cmidrule(lr){2-5}

& \multirow{2}{*}{Adj/Noun}
& LogReg  & .026 & .032 \\
& & MM-diff & \textbf{.138} & \textbf{.254} \\
\cmidrule(lr){2-5}

& \multirow{2}{*}{Noun/Adj}
& LogReg  & .028 & .040 \\
& & MM-diff & \textbf{.230} & \textbf{.666} \\

\midrule

\multirow{2}{*}{Task 2}
& \multirow{2}{*}{Noun/Noun}
& LogReg  & .011 & .014 \\
& & MM-diff & \textbf{.355} & \textbf{.564} \\

\bottomrule
\end{tabular}

\caption{Normalized indirect effects (NIEs) of mass-mean difference (MM-diff) and logistic regression (LogReg) probes on the \textit{identification} and \textit{interpretation} tasks for LLaMA-3.1-8B. Higher NIE values indicate stronger causal steering effects.}
\label{tab:nie_llama}
\end{table}

\subsection{Model steering}
\subsubsection{Identification and Interpretation}
On the identification and interpretation tasks on \textbf{NOUN} and \textbf{ADJ}, we intervene on model outputs as in Eq. \eqref{eq:mm}.
Then, we construct a linear probe using the direction $\boldsymbol{\theta} = \boldsymbol{\mu_{\text{met}}}-\boldsymbol{\mu_{\text{lit}}}$:
\[
p(\text{metaphor} \mid s) = \sigma\big(\boldsymbol{\theta}^\top \mathbf{h}(s)\big),
\]
where $\sigma(x) = \frac{1}{1 + e^{-x}}$.
For the \textit{interpretation} task, we define directions for apt/inapt pairs using the corresponding means.

First, we use the mass-mean probe to flip the true positive outputs in test set into the opposite label. We then compare MM probes with standard logistic regression (LR) probes trained on the same representations. For both probes, we calculate the normalized indirect effect (NIE) of the intervention. For details of the NIE calculation, see Appendix \ref{sec:NIEs}. Results for Llama-3.1-8B are reported in Table~\ref{tab:nie_llama}. Results across all models are in Appendix \ref{sec:NIEs}.

\subsubsection{Mass-mean and DAS steering}

We compare the performance of mass-mean steering and subspace steering using the {global} DAS matrices trained in Section \ref{sec:DAS_training}. We choose initially incorrect predictions on the identification task on the \textbf{DOM} dataset. We use the two steering methods to steer output to the correct prediction. Results are reported in Table \ref{tab:flip-rates-short}. We see that we are able to achieve fairly high performance across models.  For a full overview, see Table \ref{tab:flip-rates} in Appendix \ref{sec:das_steering_app}.

\begin{table}[t]
\centering
\small
\setlength{\tabcolsep}{6pt}
\renewcommand{\arraystretch}{1.05}
\begin{tabular}{lcc}
\toprule
\textbf{Model} & \textbf{MM flip rate} & \textbf{DAS flip rate} \\
\midrule
Gemma 2 9B          & 75.0\% & \textbf{86.0\%} \\
LLaMA 3.1 8B        & 54.0\% & \textbf{71.0\%} \\
LLaMA 3.1 70B       & 57.0\% & \textbf{65.0\%} \\
Mistral 7B          & \textbf{64.0\%} & 45.0\% \\
Mixtral 8$\times$7B & \textbf{72.0\%} & 61.0\% \\
Qwen 2.5 7B         & \textbf{62.0\%} & 61.0\% \\
Qwen 2.5 72B        & 88.0\% & \textbf{91.0\%} \\
\bottomrule
\end{tabular}
\caption{Repair rates on initially-incorrect items. \textbf{MM flip rate}: proportion of incorrect items flipped to the correct answer by full-residual-stream patching. \textbf{DAS flip rate}: proportion flipped by intervening only on the learned subspace. Bold marks the higher value per row.}
\label{tab:flip-rates-short}
\end{table}

\subsubsection{Generation}
Whilst there are differences between models, we find that $\alpha=5$ enables the most effective steering on average. Results for Llama-3.1-8B are presented in Table \ref{tab:flip-rates-alpha5}. For a full overview, we refer to Tables \ref{tab:literal-to-metaphor-ci} and \ref{tab:metaphor-to-literal-ci} in Appendix \ref{sec:generation_steering}.

\begin{table}[t]
\centering
\small
\setlength{\tabcolsep}{6pt}
\renewcommand{\arraystretch}{1.1}
\begin{tabular}{lcccc}
\toprule
 & \multicolumn{2}{c}{\textbf{Lit $\to$ met}} & \multicolumn{2}{c}{\textbf{Met $\to$ lit}} \\
\cmidrule(lr){2-3} \cmidrule(lr){4-5}
\textbf{Model} & \textbf{MM} & \textbf{DAS} & \textbf{MM} & \textbf{DAS} \\
\midrule
Gemma 2 9B          & 0.33 & 0.12 & 0.25 & 0.12 \\
LLaMA 3.1 8B        & 0.44 & 0.34 & 0.03 & 0.09 \\
LLaMA 3.1 70B       & 0.34 & 0.28 & 0.15 & 0.16 \\
Mistral 7B          & 0.52 & 0.19 & 0.28 & 0.10 \\
Mixtral 8$\times$7B & 0.38 & 0.16 & 0.24 & 0.11 \\
Qwen 2.5 7B         & 0.26 & 0.47 & 0.21 & 0.12 \\
Qwen 2.5 72B        & 0.29 & 0.15 & 0.08 & 0.19 \\
\bottomrule
\end{tabular}
\caption{Flip rates at $\alpha = 5$ for both intervention methods and both directions.}
\label{tab:flip-rates-alpha5}
\end{table}

To validate the results of the LLM-as-judge, we qualitatively inspect the generated outputs. We display some generations in Table \ref{tab:generation-examples}. More examples, including successful and unsuccessful generations, are included in 
Table \ref{tab:steering-examples} in the Appendix.

\section{Discussion and Conclusions}
Our results show that there are clear directions and subspaces associated with metaphoricity across a range of models and datasets (Figure \ref{fig:pcaplots_llama} and Figures \ref{fig:pca-adj}, \ref{fig:pca-noun} in the Appendix). We are able to train DAS matrices which effectively alter output on a classification task (Table \ref{tab:global-das-iia}), and we see that training a global metaphor subspace is more effective than training domain-specific subspaces (Table \ref{tab:das_domain_llama}), indicating that we have learnt a general concept of metaphor. These directions are moreover causally effective: using the mass-mean probe we are able to steer output in classification tasks, and this steering is more effective than steering using the direction identified by logistic regression (Table \ref{tab:nie_llama}). We do note that the NIE values are somewhat lower than in \cite{marks2024geometry}, however, it is also the case that metaphor identification and interpretation is a more difficult task than judging truth and falsity. We also see that there are higher NIEs steering from literal to metaphor than vice-versa. Creating a literal meaning from a metaphorical meaning is perhaps conceptually more difficult: \textit{warm coffee} can readily be interpreted metaphorically via personification of the coffee, whereas \textit{warm welcome} is more difficult to interpret literally.

\begin{table}[htbp]
\centering
\caption{Example generations for LLaMA-8B under steering at $\alpha=5$. Baseline = unintervened model, MM = mass-mean steering, DAS = DAS intervention. Both examples: true label \emph{literal}, target \emph{metaphorical}.}
\label{tab:generation-examples}
\small
\resizebox{\columnwidth}{!}{%
\begin{tabular}{@{}l l p{4cm}@{}}
\toprule
Method & Judge prediction & Generation \\
\midrule
\multicolumn{3}{@{}l}{\textit{adj-noun pair:} \texttt{icy pavement}} \\
\midrule
Baseline & literal      & extremely cold and slippery. \\
MM       & metaphorical & extremely cold and unfriendly. \\
DAS      & metaphorical & extremely cold and unfriendly. \\
\midrule
\multicolumn{3}{@{}l}{\textit{adj-noun-pair:} \texttt{heated washer}} \\
\midrule
Baseline & literal      & having a high temperature. \\
MM       & metaphorical & having a high temperature, often due to strong emotions. \\
DAS      & metaphorical & having a strong, intense, or passionate quality. \\

\bottomrule
\end{tabular}
 }
\end{table}

We further steer generation to more metaphorical output. In Table \ref{tab:flip-rates-alpha5}, we see that the mass-mean probes are generally more effective than DAS steering, although this differs across model. Example generations are given in Table \ref{tab:generation-examples}. In future work, this technique could be developed to give a way of controlling the metaphoricity of generated output in a more direct way than prompting.

We acknowledge that IIA scores are moderate and variable, and indeed this does imply that a proportion of the metaphoricity information is captured elsewhere. We note however, that previous work which has used DAS for locating concepts in model activation achieve similar scores (see e.g. \cite{mueller2025mib}). Achieving a perfect IIA score is not necessarily achievable in practice \cite{geiger2024finding}, and this is a limitation of using DAS for concept-finding.

On the basis of Table \ref{tab:das_domain_llama}, we concluded that the DAS matrix trained on all domains captures a general concept of metaphoricity, as evidenced by its stronger transfer IIA results compared to the results achieved with its domain-specific counterpart. One confound could be the training data size: Compared to the domain-specific matrices trained on \textit{n} samples, the global DAS matrix is trained on those \textit{and all the training samples from all other domains}. As a control, we train a matched-n global DAS matrix. This matrix trained on the number of samples in each domain corresponding to the number of samples from the smallest training domain ($n=255$). We see in Appendix Table \ref{tab:matched-n} that for Gemma 2 9B, training a matched-n global DAS matrix still achieves a better or equal transfer IIA score in 6/8 transfer domains. 

For our generation experiment, we carried out a human validation of the LLM-as-judge. For the Gemma 9B model outputs, one of the authors of this paper has annotated 15\% (100 items) as either metaphor, literal, or incorrect (nonsensical or incorrect output). Raw agreement is 0.79, Cohen’s kappa is 0.60, indicating moderate to substantial agreement. This indicates that the judge annotations are mostly reliable. This suggests that the concept we have isolated in the classification task is at least partially sufficient to flip free-form generation. 
However, generally, we do achieve substantially lower IIA scores for the generation task compared to the classification task. Our DAS concept allows us to steer model classification to a moderately high degree, but this effect is less amplified for generation, and requires that we scale this signal by $\alpha = 5$. There can be different reasons for this. It is possible that the metaphoricity concept we isolate in the classification task is functionally adapted to this task. The model \textit{could} have another metaphoricity concept located in a different position adapted for generation. On the other hand, the fact that the concept has some meaning-steering effect suggests that we have captured a common concept of metaphoricity, at least partially.

Further, we employed a controlled, templated setup for measuring metaphoricity in LLMs. We made this choice to effectively isolate the parts of the prompt that have a causal effect on the output. The structured sentence patterns allowed us to localize causal effects to specific token positions. Therefore, we deliberately did not test representations of metaphoricity in a more naturalistic context (e.g. poems), limiting the external validity of our findings.

\subsection{Conclusions}
We have given evidence that metaphoricity is encoded as an interpretable direction in a variety of LLMs, illustrated as linear separability in PCA plots. We intervene on model activations and show that these directions are causally effective in changing model outputs in both classification and generation tasks. Future work in this area could extend to other aspects of writing style: at present, much work is based in prompting, and these techniques may give more control over style modifications.\\

\section{Limitations and Risks}
\begin{itemize}
    \item This work is carried out only on English metaphor. Linear separability and interpretable directions could be very different in other languages.
    \item Mechanistic interpretability work requires very controlled settings and datasets, whereas metaphor can extend across multiple sentences, paragraphs, and discourse. The datasets we use are therefore only very simplistic uses of metaphor. More complex metaphor use is an area for future research.
    \item We cannot carry out our experiments on the most highly performing closed models, since they do not have open weights, and computational resources would preclude this.
    \item \textbf{Risks} In its present form, risks resulting from this work are fairly low. However, model steering techniques can potentially be risky. It is possible to steer towards undesirable as well as desirable behaviour.
\end{itemize}

\section*{Acknowledgements}
This work was carried out using the National Supercomputer Snellius made available via SURF (www.surf.nl) and the University of Amsterdam. ML gratefully acknowledges financial support from the Sectorplan science fund of the Dutch government.

\bibliography{custom}
\newpage
\appendix
\section{\textbf{NOUN} identification prompt}
\label{sec:noun_prompt}
The politician is glass. This sentence is: metaphoric \\
The politician is transparent. This sentence is: literal \\
The cake is heaven. This sentence is: metaphoric \\
The cake is delicious. This sentence is: literal \\
\texttt
{sentence}. This sentence is:

\onecolumn

\clearpage
\section{Activation patching plots}
\begin{figure*}[!htbp]
\centering

\begin{subfigure}{0.23\textwidth}
\includegraphics[width=\linewidth]{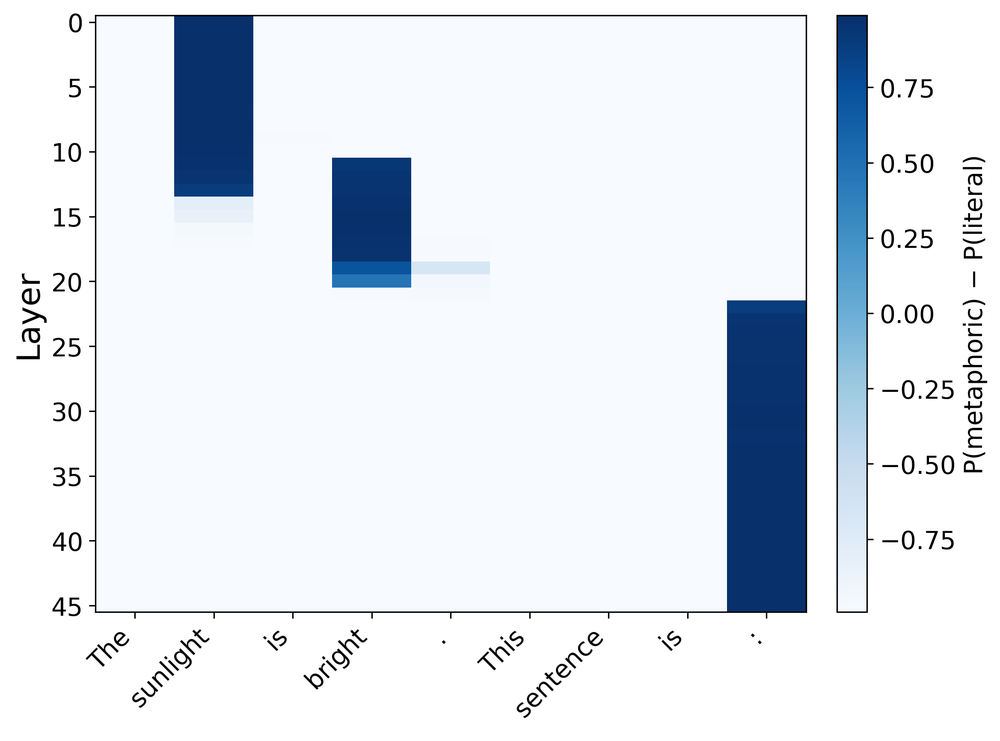}
\caption{Gemma 27B Adj}
\end{subfigure}
\hfill
\begin{subfigure}{0.23\textwidth}
\includegraphics[width=\linewidth]{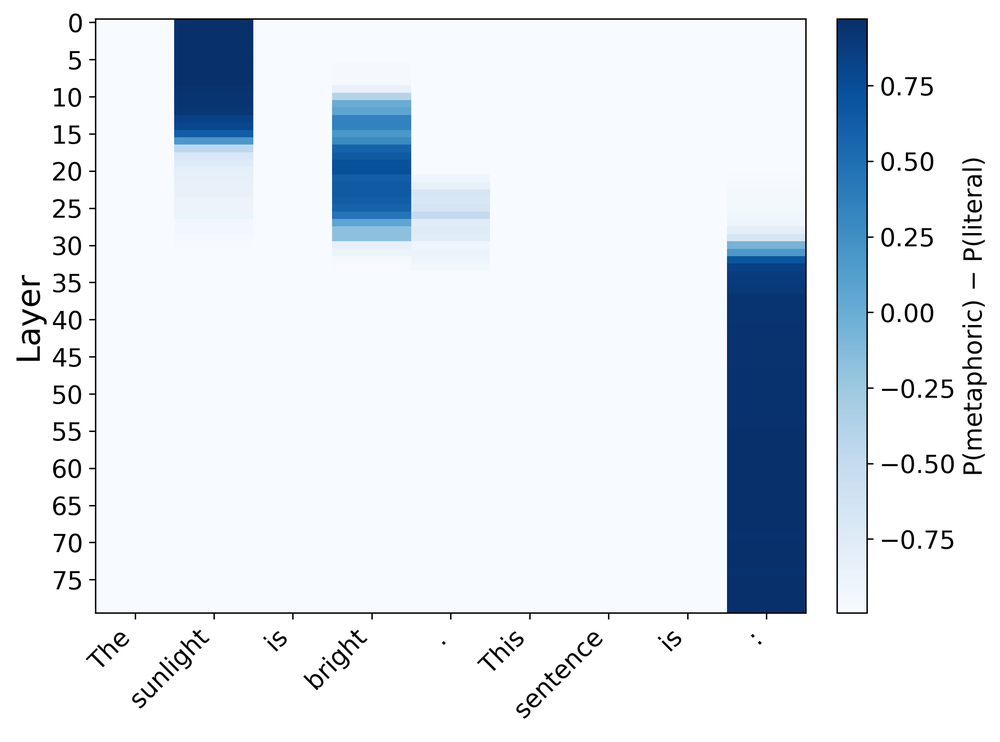}
\caption{Llama 70B Adj}
\end{subfigure}
\hfill
\begin{subfigure}{0.23\textwidth}
\includegraphics[width=\linewidth]{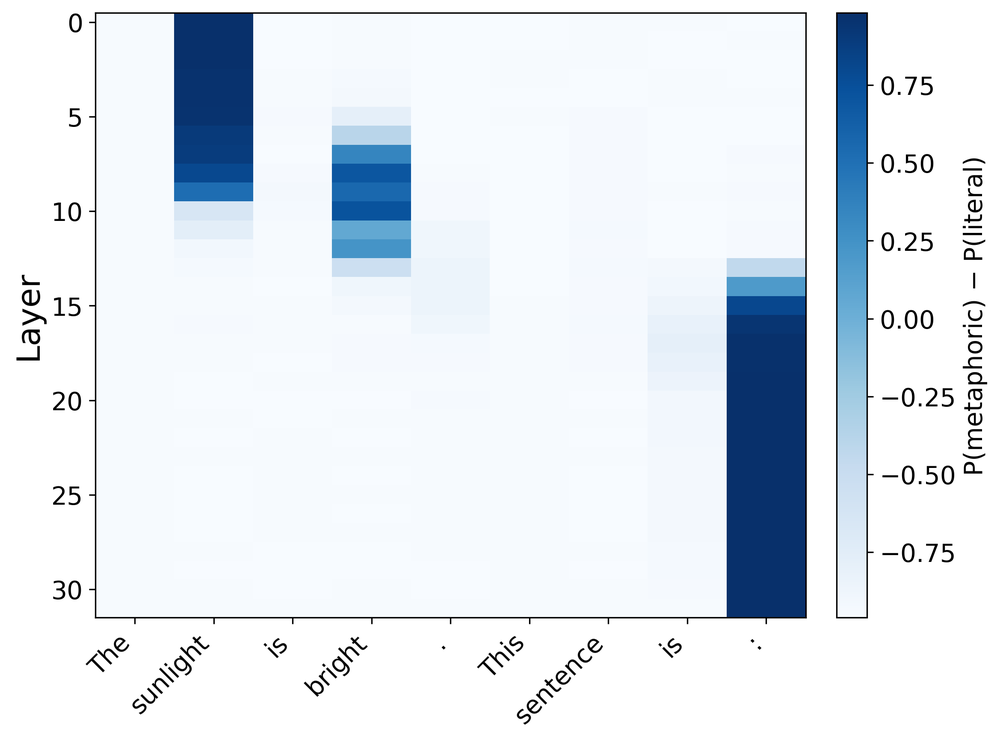}
\caption{Mixtral 8x7B Adj}
\end{subfigure}
\hfill
\begin{subfigure}{0.23\textwidth}
\includegraphics[width=\linewidth]{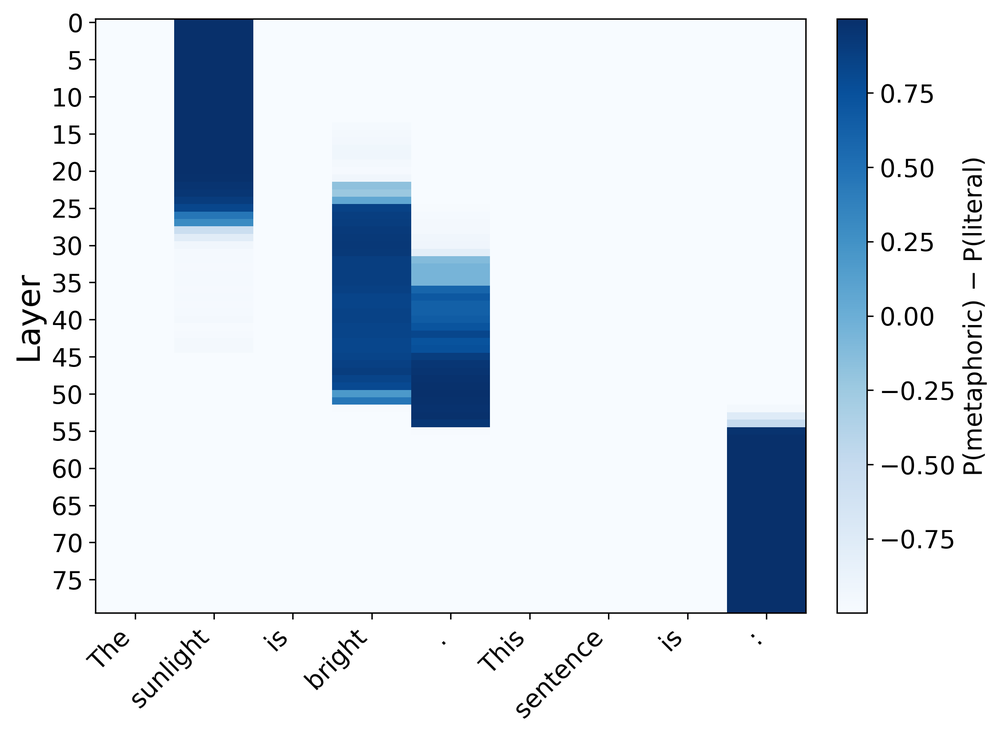}
\caption{Qwen 72B Adj}
\end{subfigure}

\vspace{0.5em}

\begin{subfigure}{0.23\textwidth}
\includegraphics[width=\linewidth]{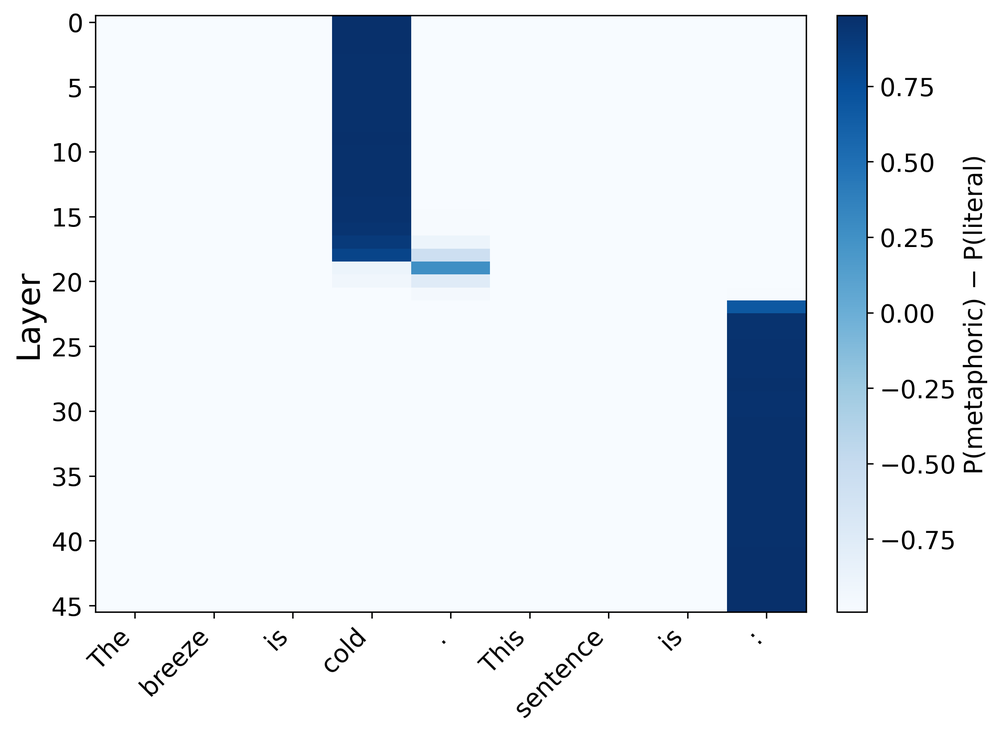}
\caption{Gemma 27B Noun}
\end{subfigure}
\hfill
\begin{subfigure}{0.23\textwidth}
\includegraphics[width=\linewidth]{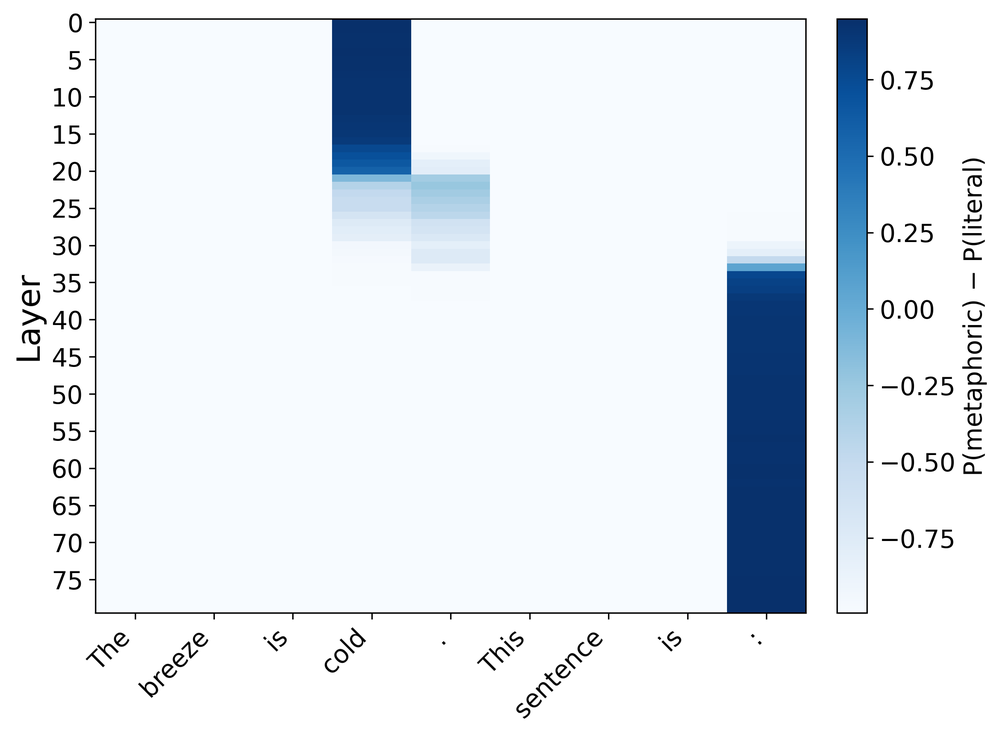}
\caption{Llama 70B Noun}
\end{subfigure}
\hfill
\begin{subfigure}{0.23\textwidth}
\includegraphics[width=\linewidth]{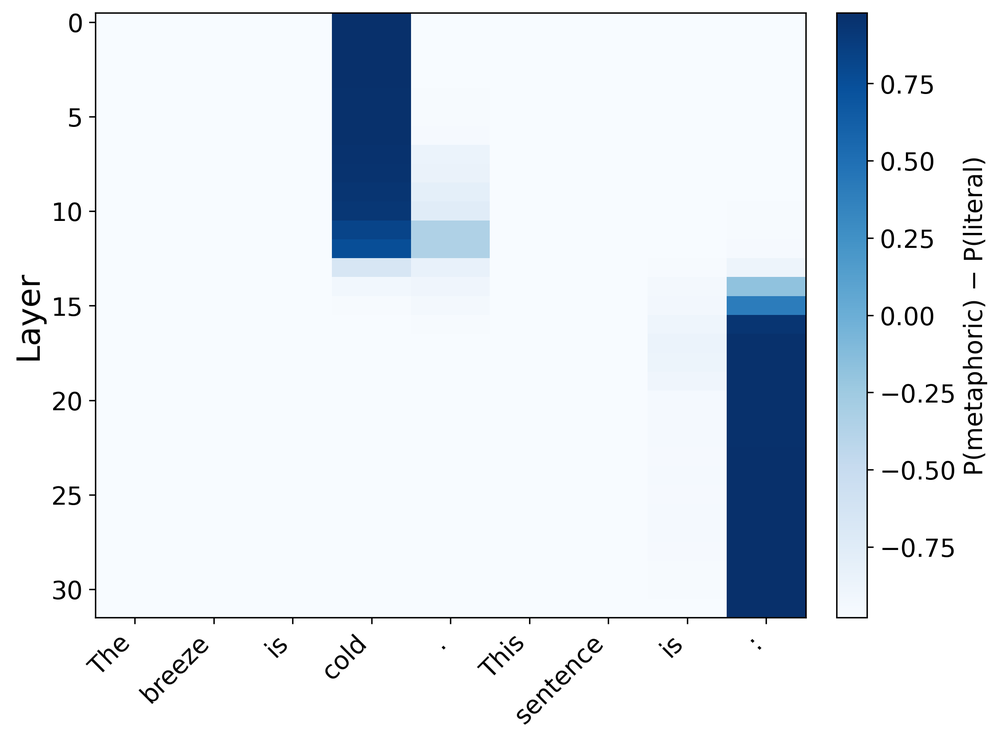}
\caption{Mixtral 8x7B Noun}
\end{subfigure}
\hfill
\begin{subfigure}{0.23\textwidth}
\includegraphics[width=\linewidth]{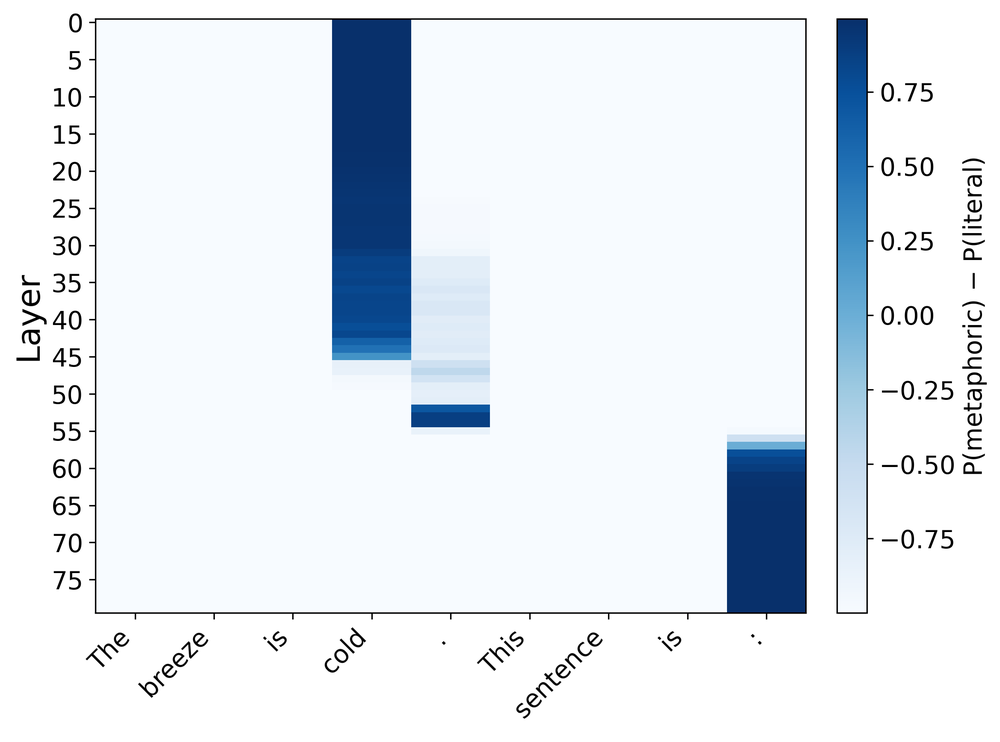}
\caption{Qwen 72B Noun}
\end{subfigure}

\caption{\textbf{Activation patching results for adjective and noun representations in the \textit{identification} task.} Activations from the metaphorical prompt are patched into the literal prompt. 
\textbf{(a)–(d) Adjective dataset.} The prompts use the literal sentence \textit{“The sunlight is bright.”} and the metaphorical sentence \textit{“The idea is bright.”}. 
\textbf{(e)–(h) Noun dataset.} The prompts use the literal sentence \textit{“The breeze is cold.”} and the metaphorical sentence \textit{“The breeze is ice.”}.}
\label{fig:patching-task1}
\end{figure*}

\begin{figure*}[!htbp]
\begin{subfigure}{0.32\textwidth}
\includegraphics[width=\linewidth]{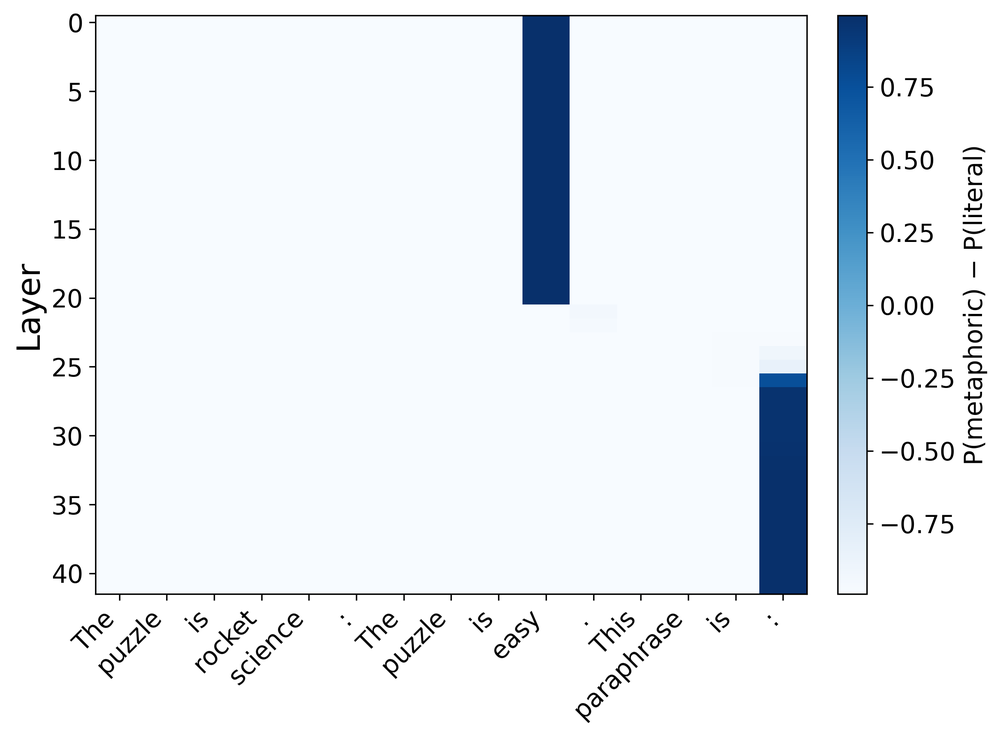}
\caption{Gemma 9B Noun}
\end{subfigure}
\hfill
\begin{subfigure}{0.32\textwidth}
\includegraphics[width=\linewidth]{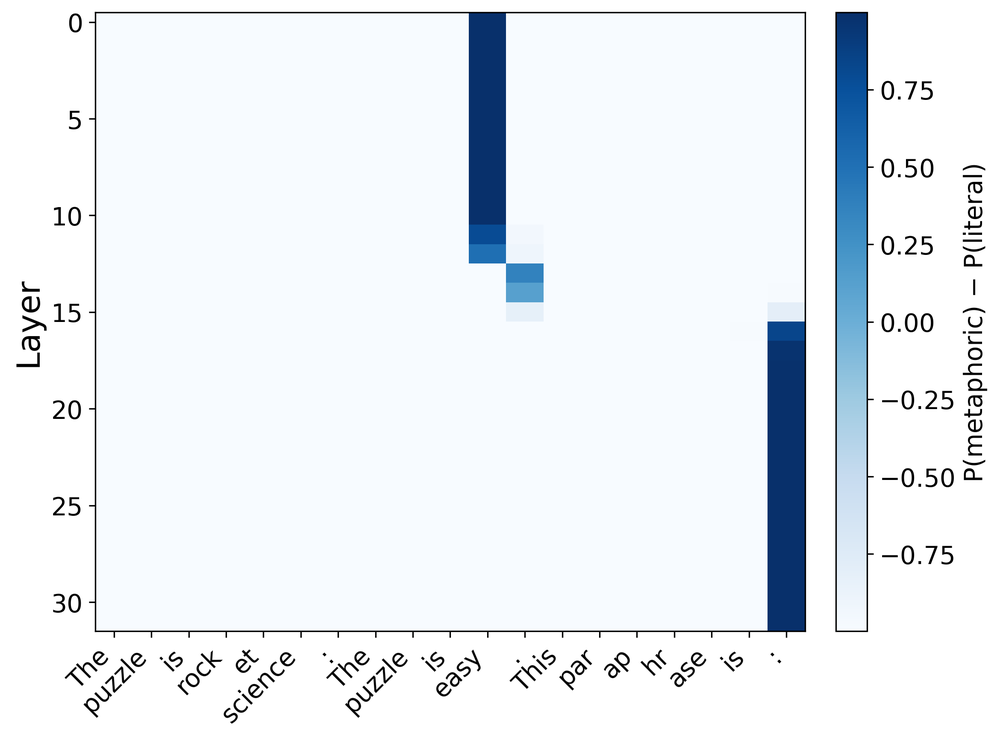}
\caption{Mistral 7B Noun}
\end{subfigure}
\hfill
\begin{subfigure}{0.32\textwidth}
\includegraphics[width=\linewidth]{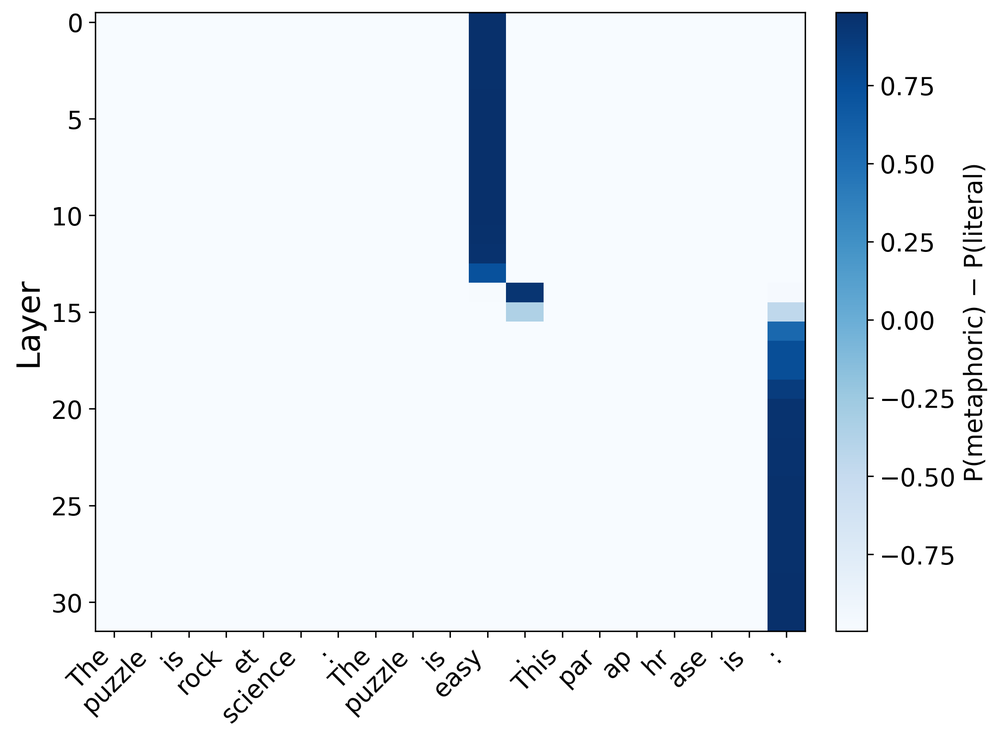}
\caption{Mixtral 8x7B Noun}
\end{subfigure}
\caption{\textbf{Activation patching results for noun representations in the \textit{interpretation} task.} 
\label{fig:patching-task2}
Activations from the inapt paraphrasing prompt are patched into the apt paraphrasing prompt. The prompts use the metaphorical sentence \textit{“The puzzle is rocket science.”}, the correct paraphrase \textit{“The puzzle is difficult.”}, and the incorrect paraphrase \textit{“The puzzle is easy.”}.}
\end{figure*}

\clearpage

\section{PCA plots}
\label{sec:PCAplots_allmodels}
\begin{figure*}[!htbp]
\centering

\begin{subfigure}{0.23\textwidth}
\includegraphics[width=\linewidth]{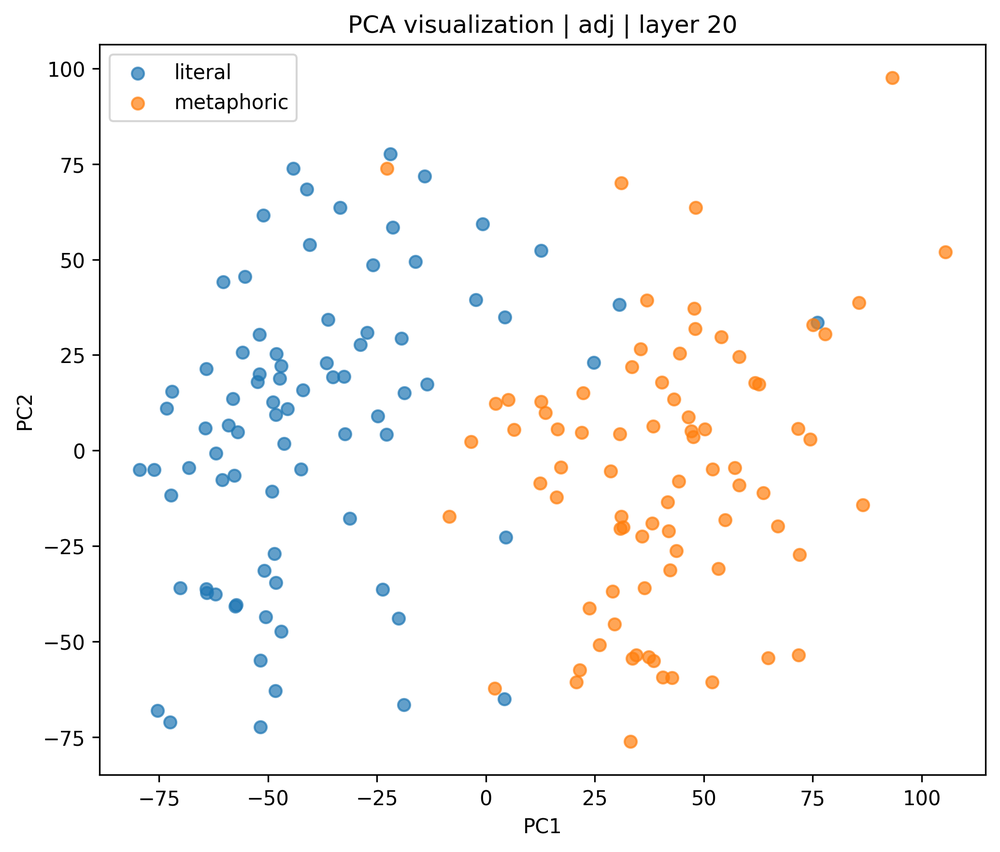}
\caption{Gemma-2 9B}
\end{subfigure}
\hfill
\begin{subfigure}{0.23\textwidth}
\includegraphics[width=\linewidth]{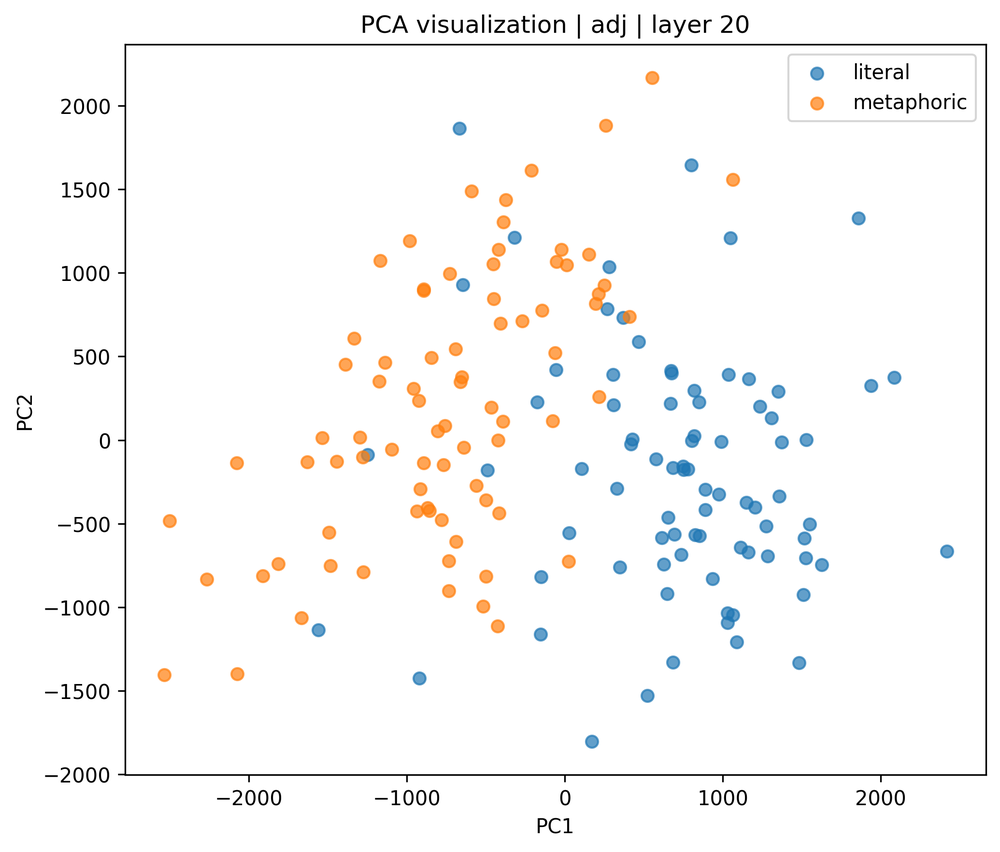}
\caption{Gemma-2 27B}
\end{subfigure}
\hfill
\begin{subfigure}{0.23\textwidth}
\includegraphics[width=\linewidth]{pic/pca/task1/pca_adj_layer11_meta-llama_Llama-3.1-8B-Instruct.png}
\caption{Llama-3.1 8B}
\end{subfigure}
\hfill
\begin{subfigure}{0.23\textwidth}
\includegraphics[width=\linewidth]{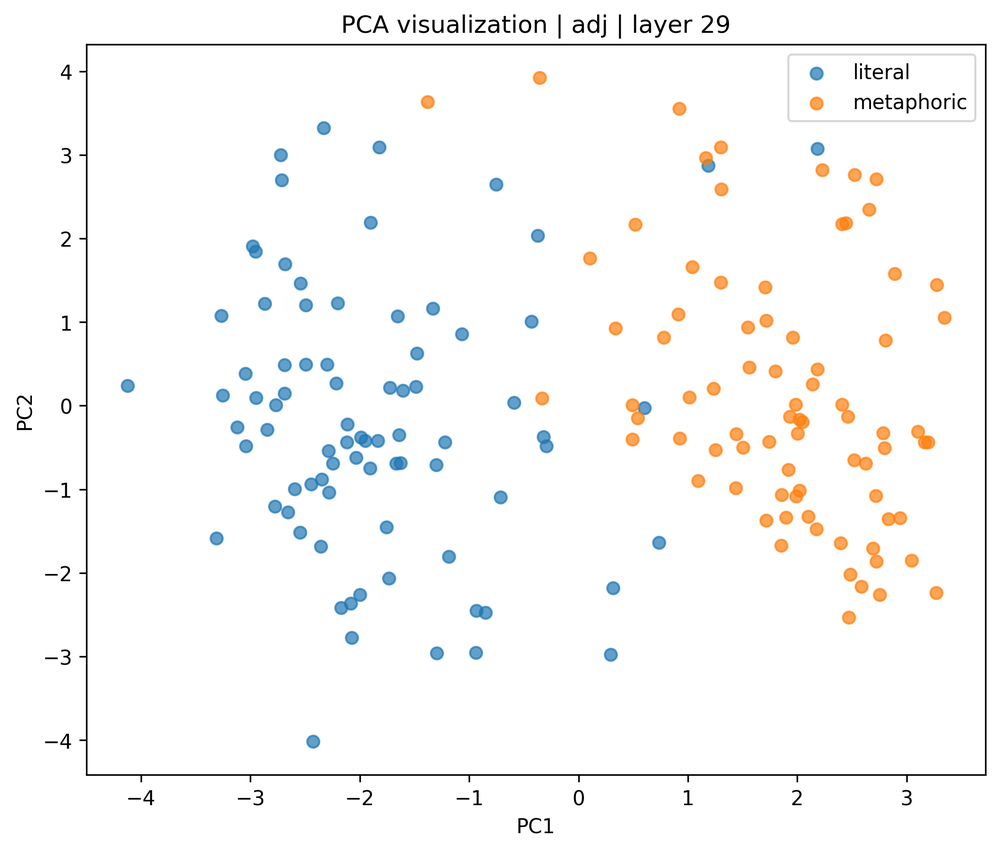}
\caption{Llama-3.1 70B}
\end{subfigure}

\vspace{0.5em}

\begin{subfigure}{0.23\textwidth}
\includegraphics[width=\linewidth]{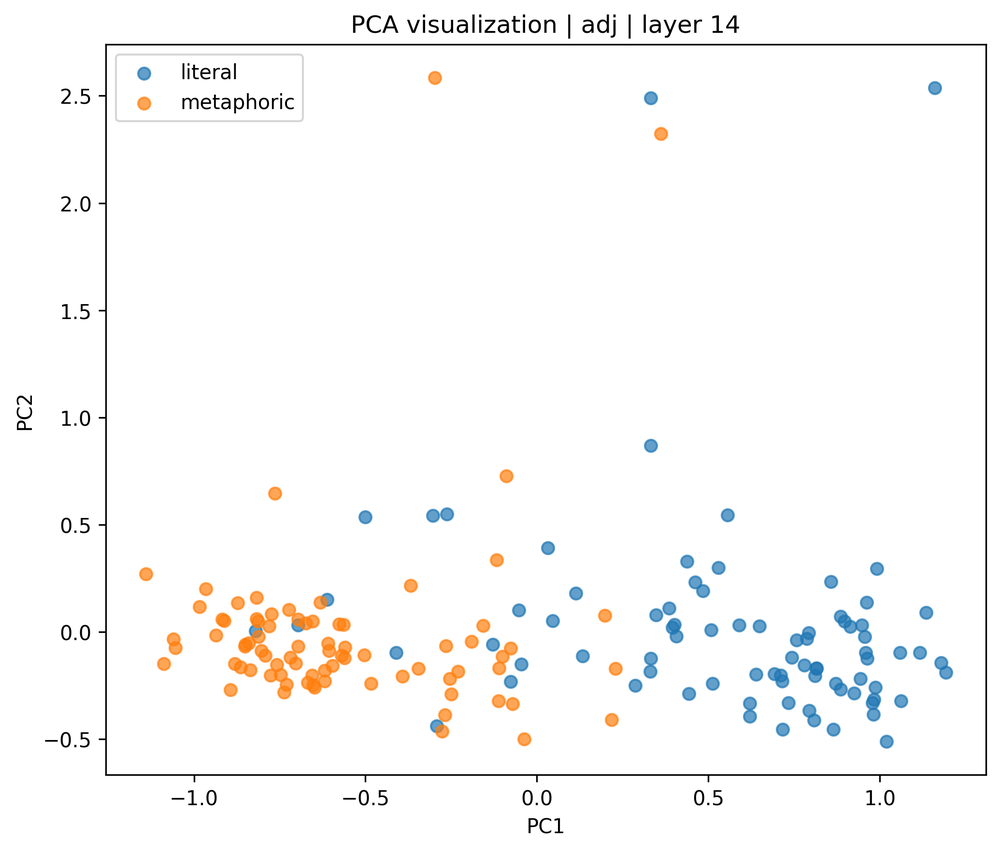}
\caption{Mistral 7B}
\end{subfigure}
\hfill
\begin{subfigure}{0.23\textwidth}
\includegraphics[width=\linewidth]{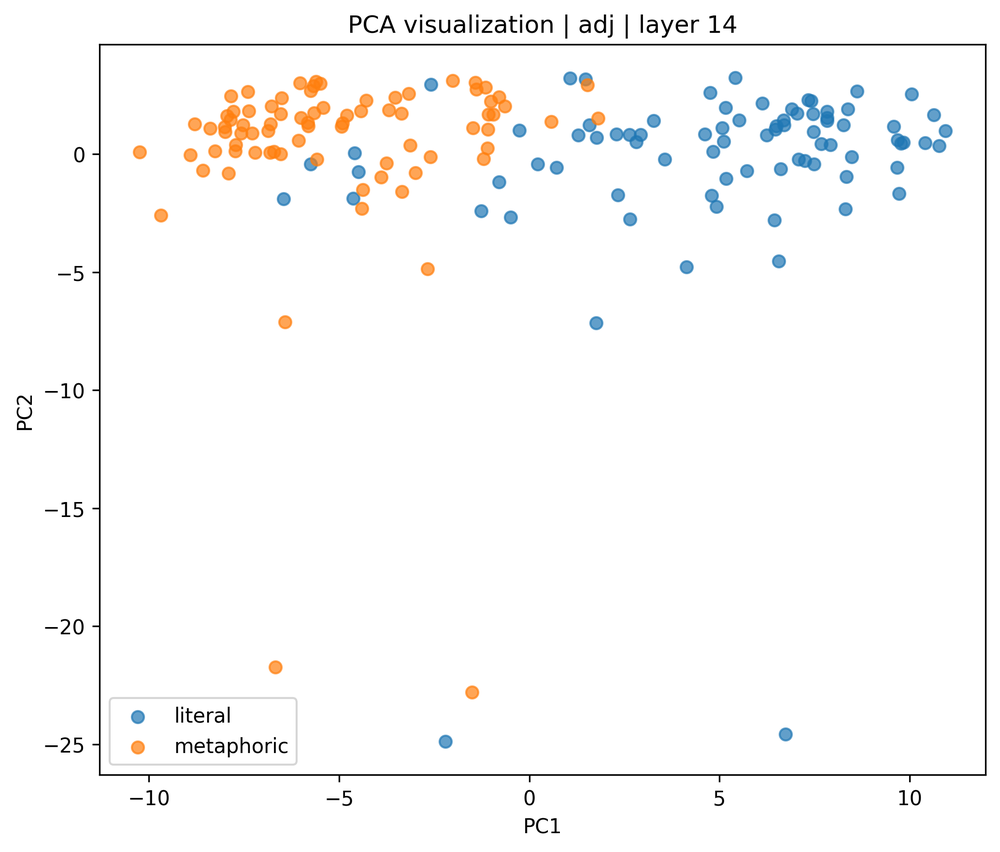}
\caption{Mixtral 8x7B}
\end{subfigure}
\hfill
\begin{subfigure}{0.23\textwidth}
\includegraphics[width=\linewidth]{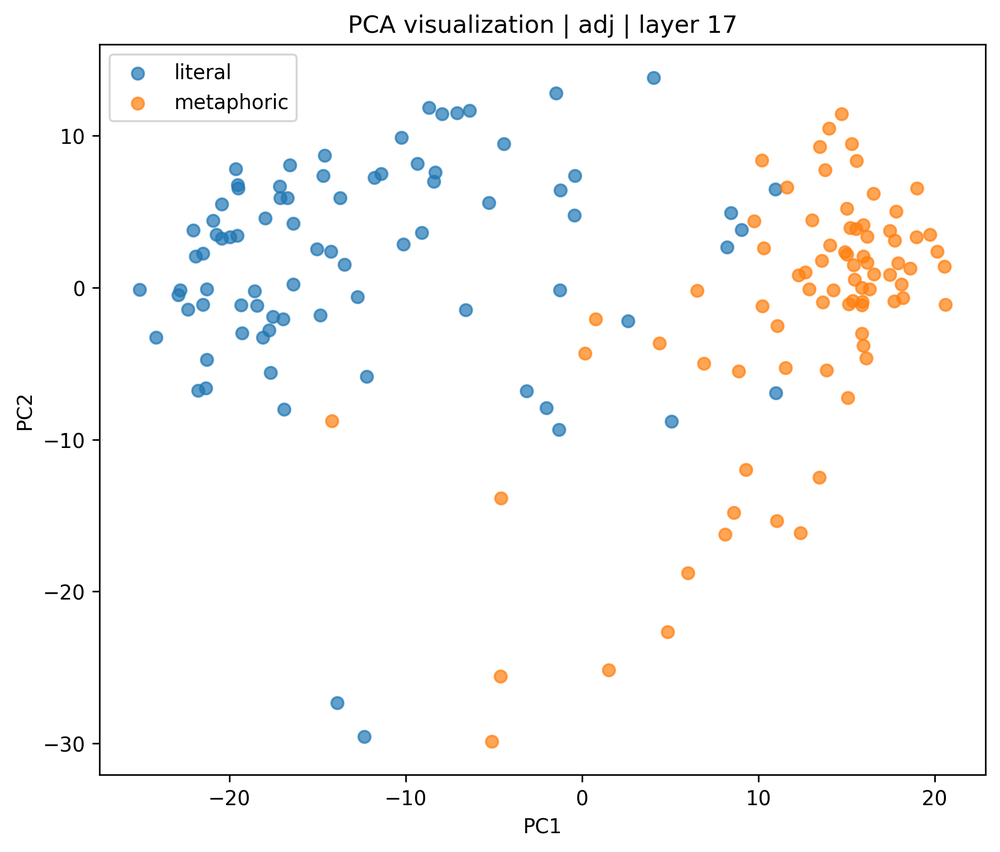}
\caption{Qwen2.5 7B}
\end{subfigure}
\hfill
\begin{subfigure}{0.23\textwidth}
\includegraphics[width=\linewidth]{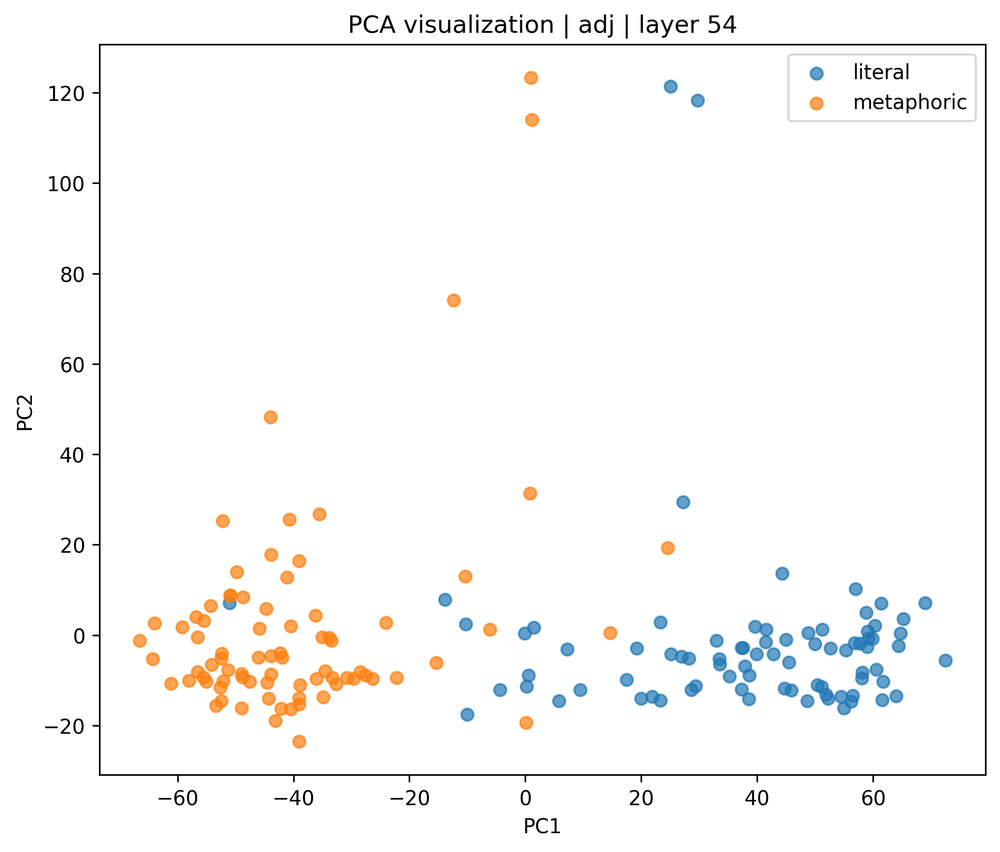}
\caption{Qwen2.5 72B}
\end{subfigure}

\caption{Identification task: PCA visualizations for adjective representations.}
\label{fig:pca-adj}
\end{figure*}

\begin{figure*}[!htbp]
\centering

\begin{subfigure}{0.23\textwidth}
\includegraphics[width=\linewidth]{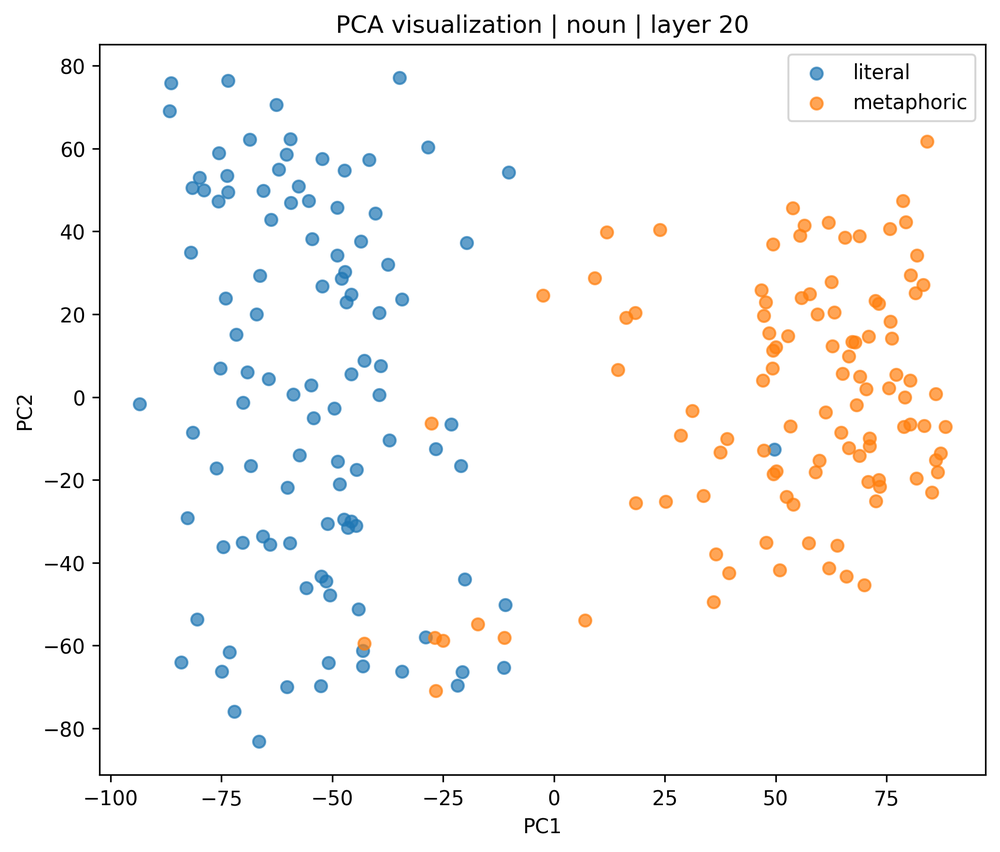}
\caption{Gemma-2 9B}
\end{subfigure}
\hfill
\begin{subfigure}{0.23\textwidth}
\includegraphics[width=\linewidth]{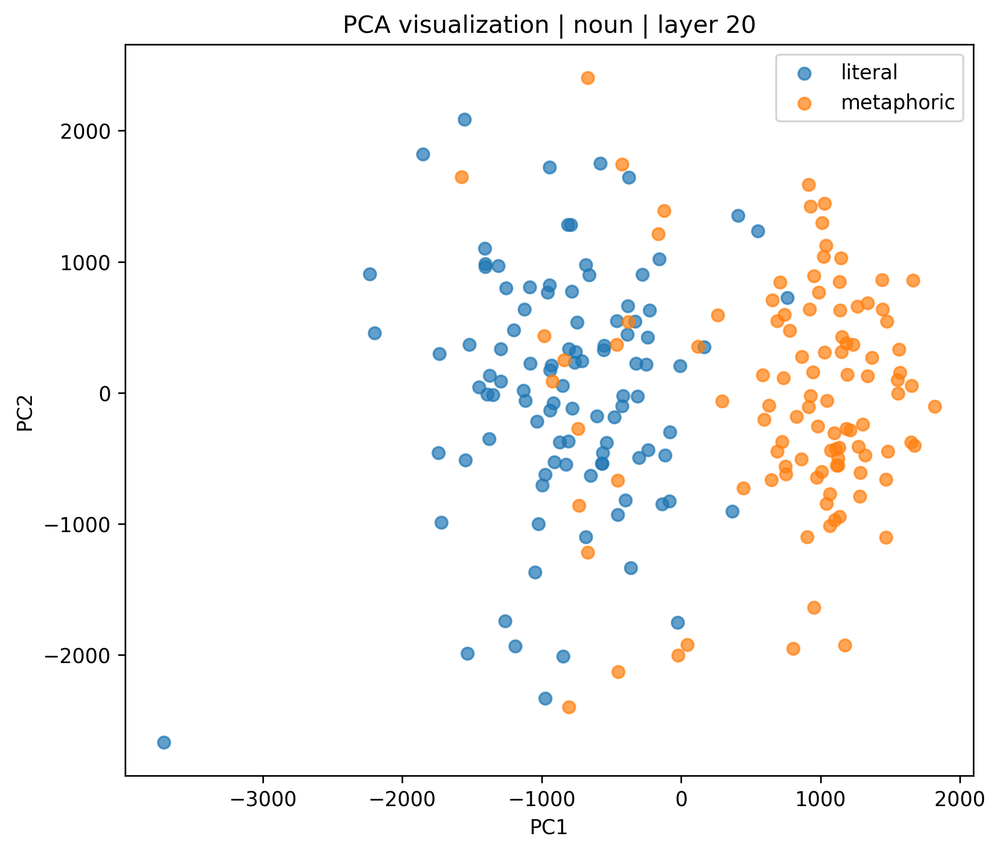}
\caption{Gemma-2 27B}
\end{subfigure}
\hfill
\begin{subfigure}{0.23\textwidth}
\includegraphics[width=\linewidth]
{pic/pca/task1/pca_noun_layer11_meta-llama_Llama-3.1-8B-Instruct.png}
\caption{Llama-3.1 8B}
\end{subfigure}
\hfill
\begin{subfigure}{0.23\textwidth}
\includegraphics[width=\linewidth]{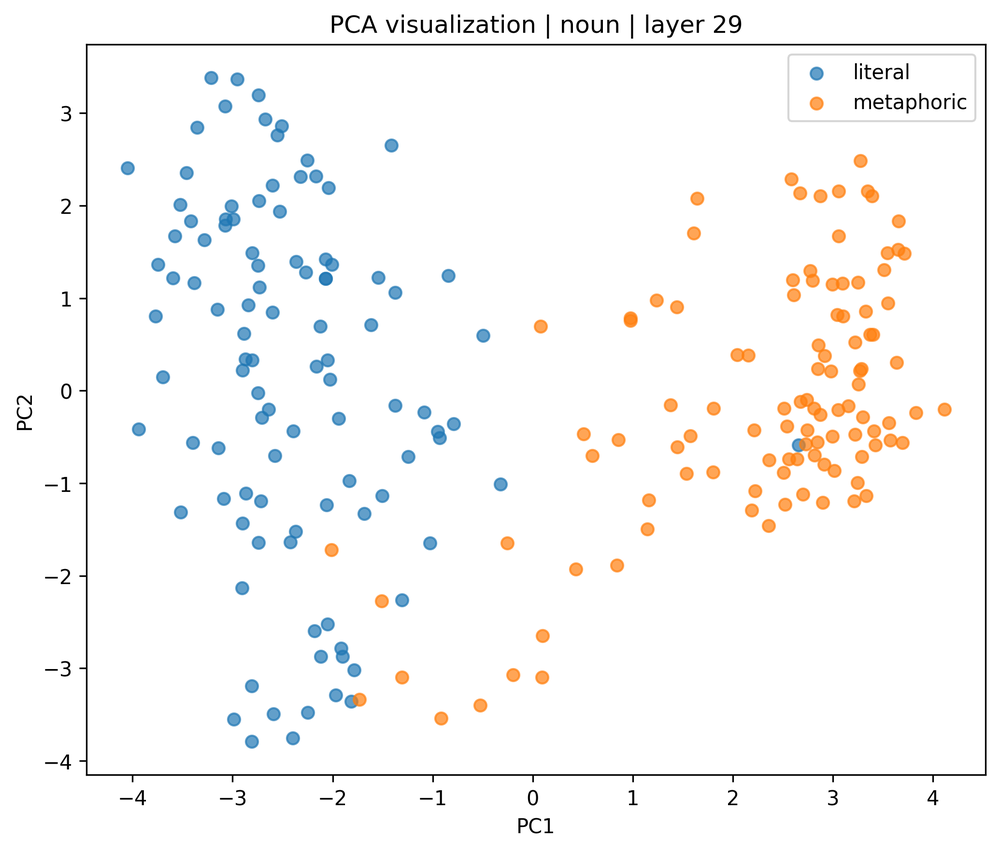}
\caption{Llama-3.1 70B}
\end{subfigure}

\vspace{0.5em}

\begin{subfigure}{0.23\textwidth}
\includegraphics[width=\linewidth]{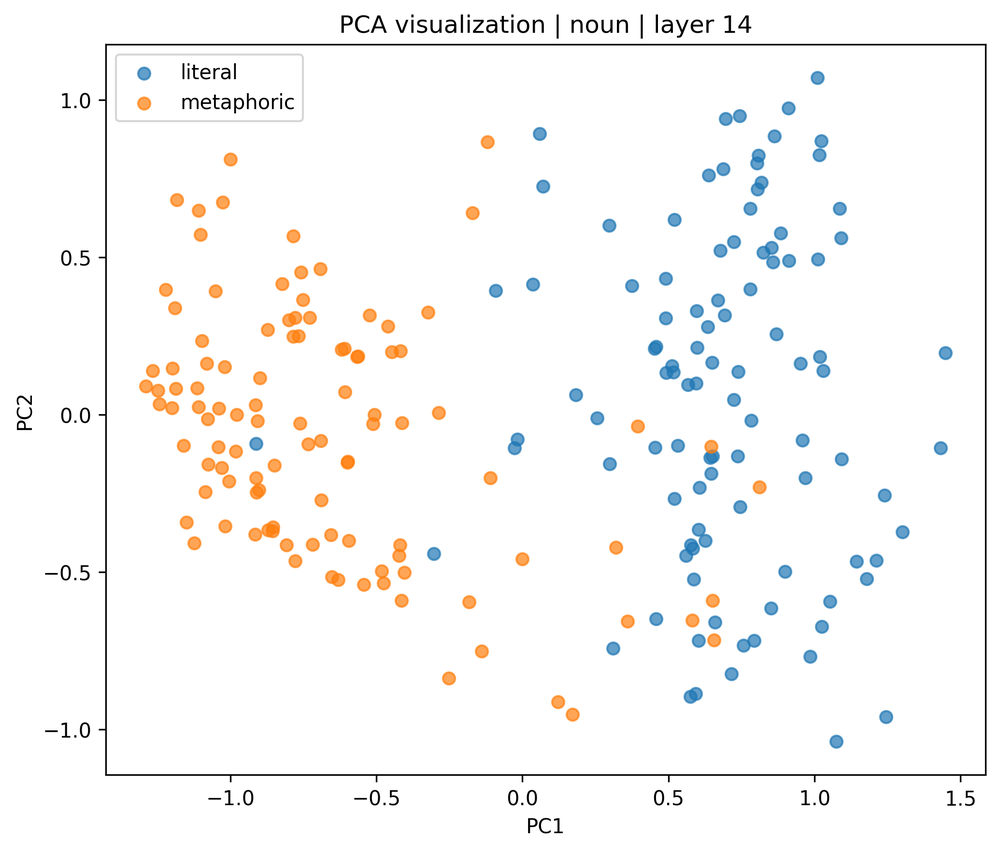}
\caption{Mistral 7B}
\end{subfigure}
\hfill
\begin{subfigure}{0.23\textwidth}
\includegraphics[width=\linewidth]{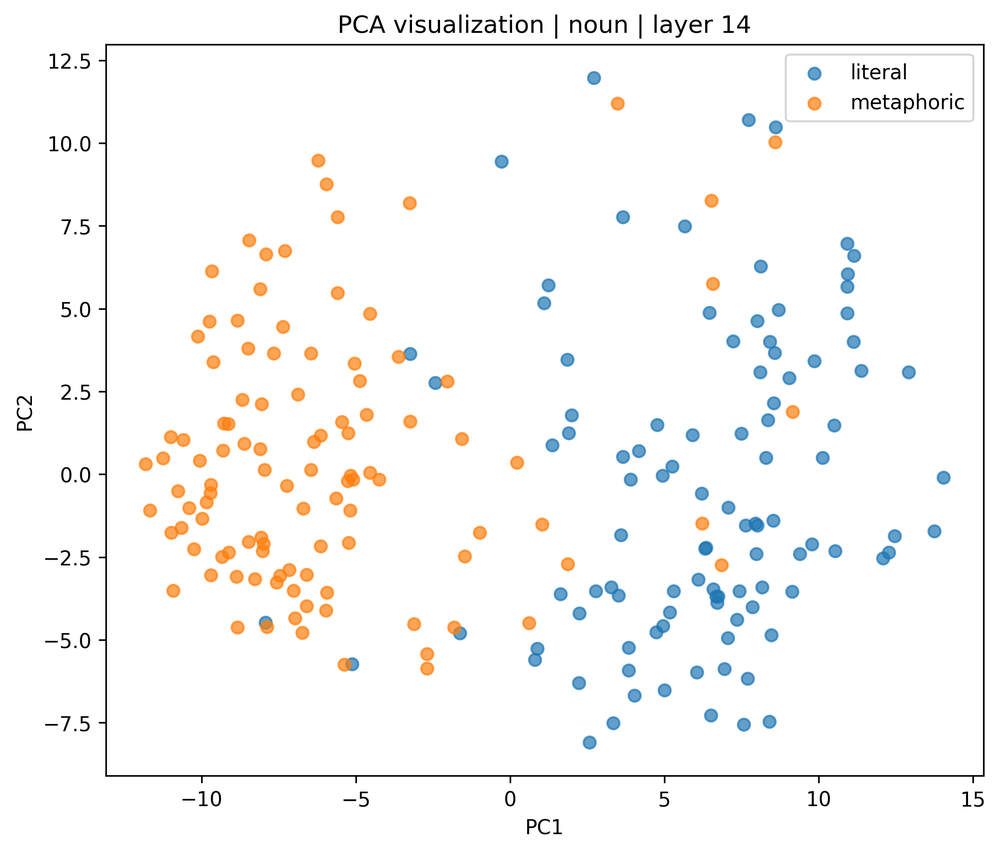}
\caption{Mixtral 8x7B}
\end{subfigure}
\hfill
\begin{subfigure}{0.23\textwidth}
\includegraphics[width=\linewidth]{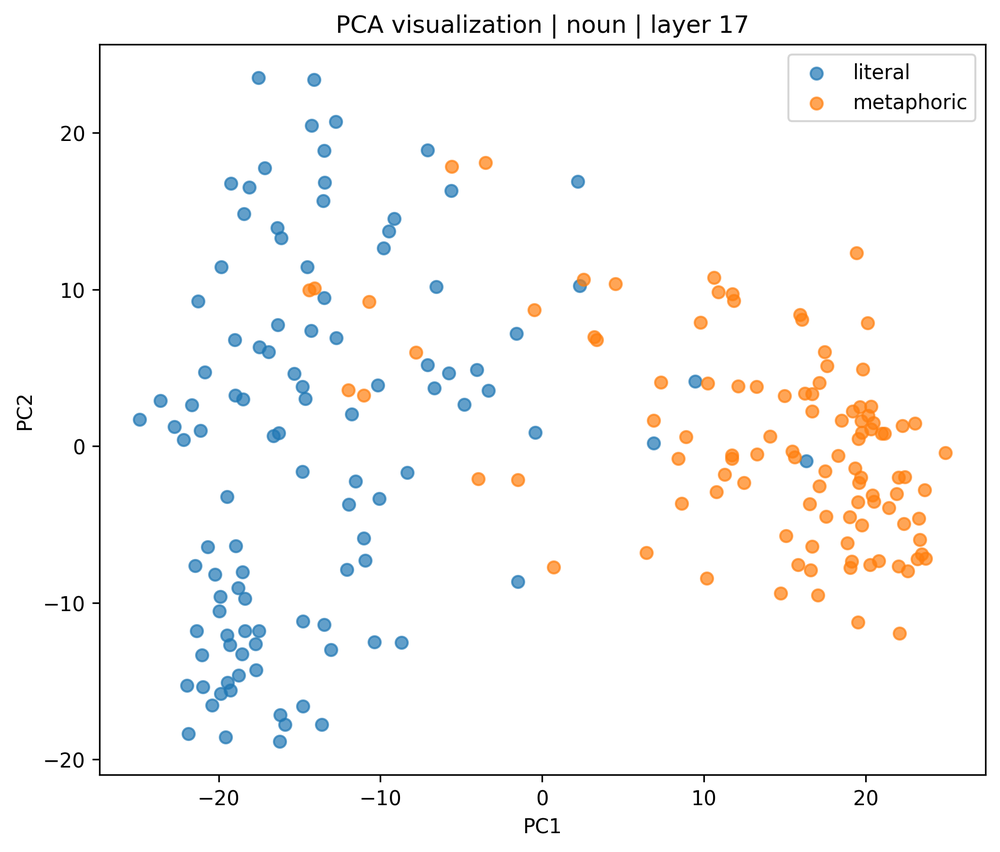}
\caption{Qwen2.5 7B}
\end{subfigure}
\hfill
\begin{subfigure}{0.23\textwidth}
\includegraphics[width=\linewidth]{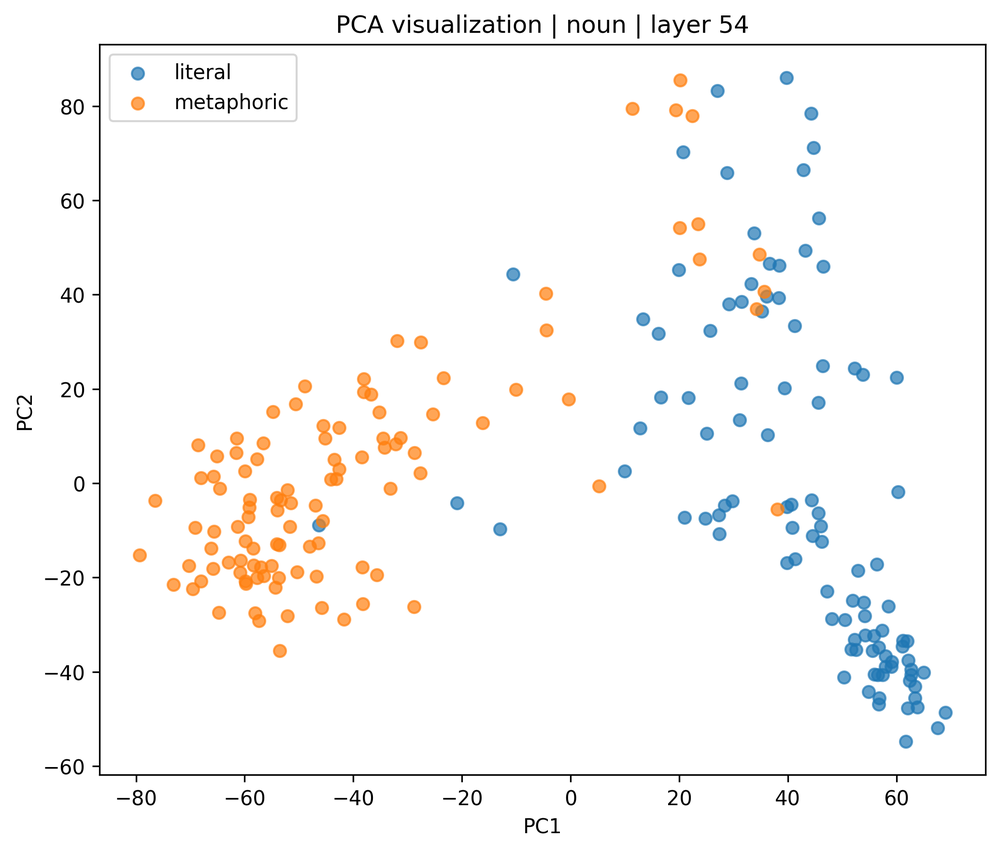}
\caption{Qwen2.5 72B}
\end{subfigure}

\caption{Identification task: PCA visualizations for noun representations.}
\label{fig:pca-noun}
\end{figure*}

\begin{figure*}[h!]
\centering

\begin{subfigure}{0.24\textwidth}
\includegraphics[width=\linewidth]{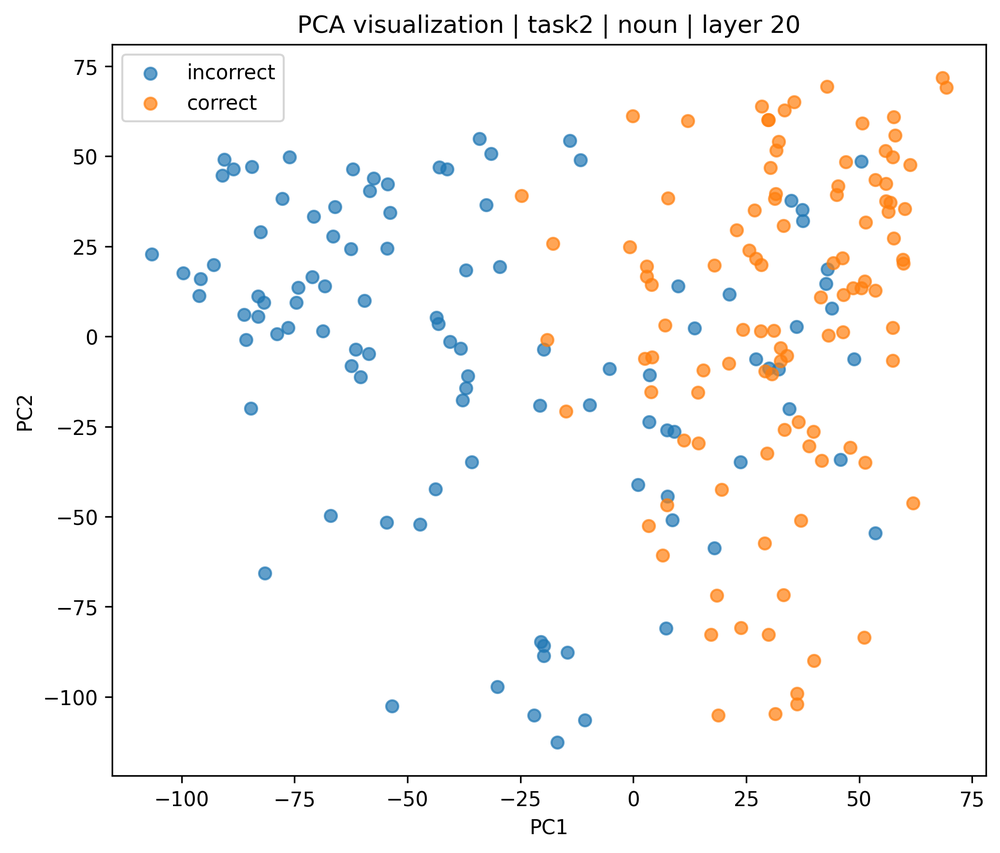}
\caption{Gemma-2 9B}
\end{subfigure}
\hfill
\begin{subfigure}{0.24\textwidth}
\includegraphics[width=\linewidth]{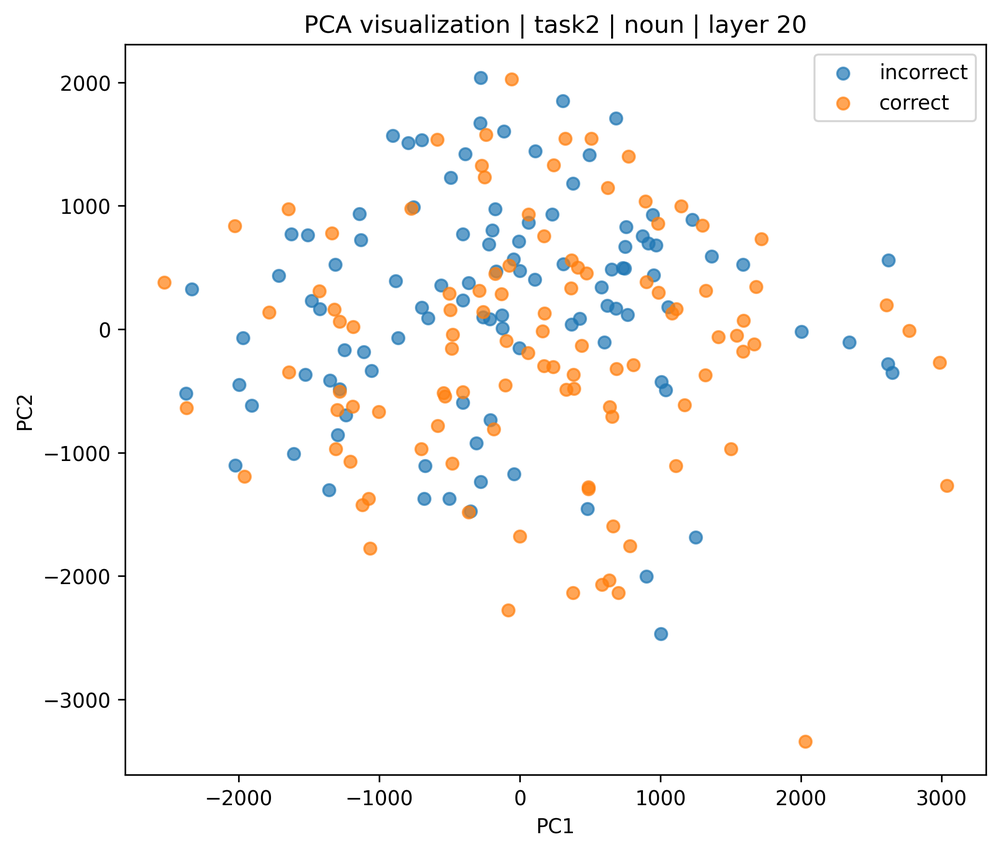}
\caption{Gemma-2 27B}
\end{subfigure}
\hfill
\begin{subfigure}{0.24\textwidth}
\includegraphics[width=\linewidth]{pic/pca/task2/pca_task2_noun_layer11_meta-llama_Llama-3.1-8B-Instruct.png}
\caption{Llama-3.1 8B}
\end{subfigure}
\hfill
\begin{subfigure}{0.24\textwidth}
\includegraphics[width=\linewidth]{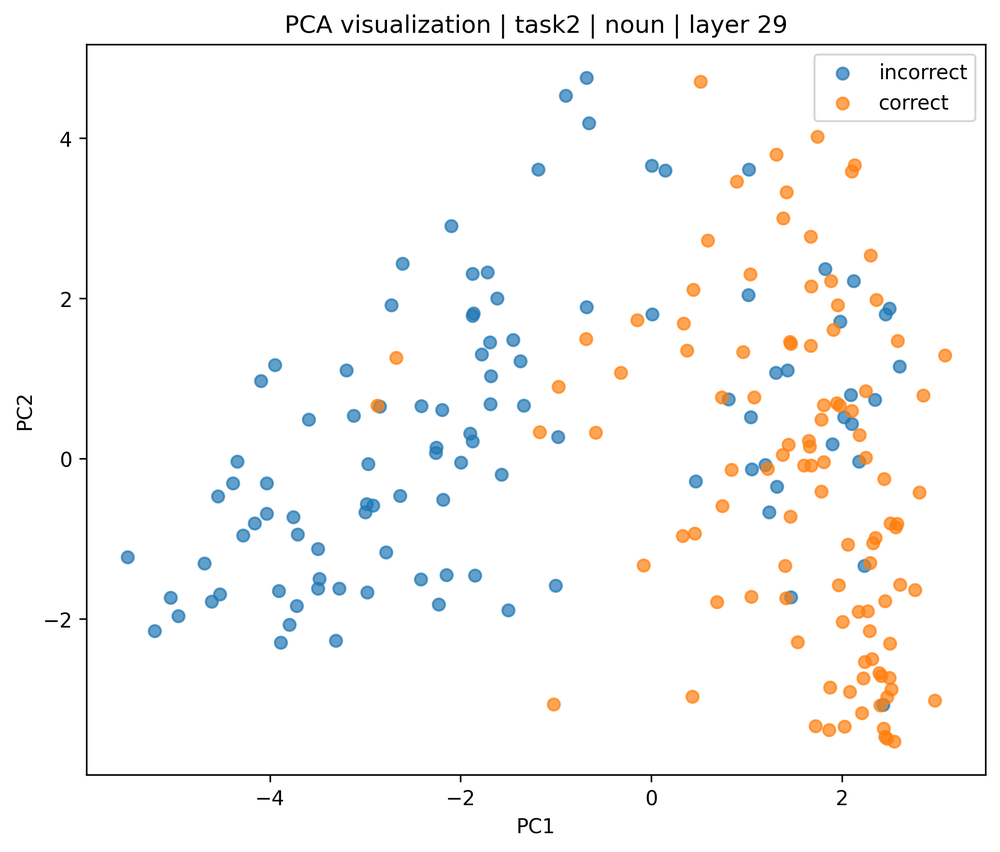}
\caption{Llama-3.1 8B}
\end{subfigure}
\hfill
\begin{subfigure}{0.24\textwidth}
\includegraphics[width=\linewidth]{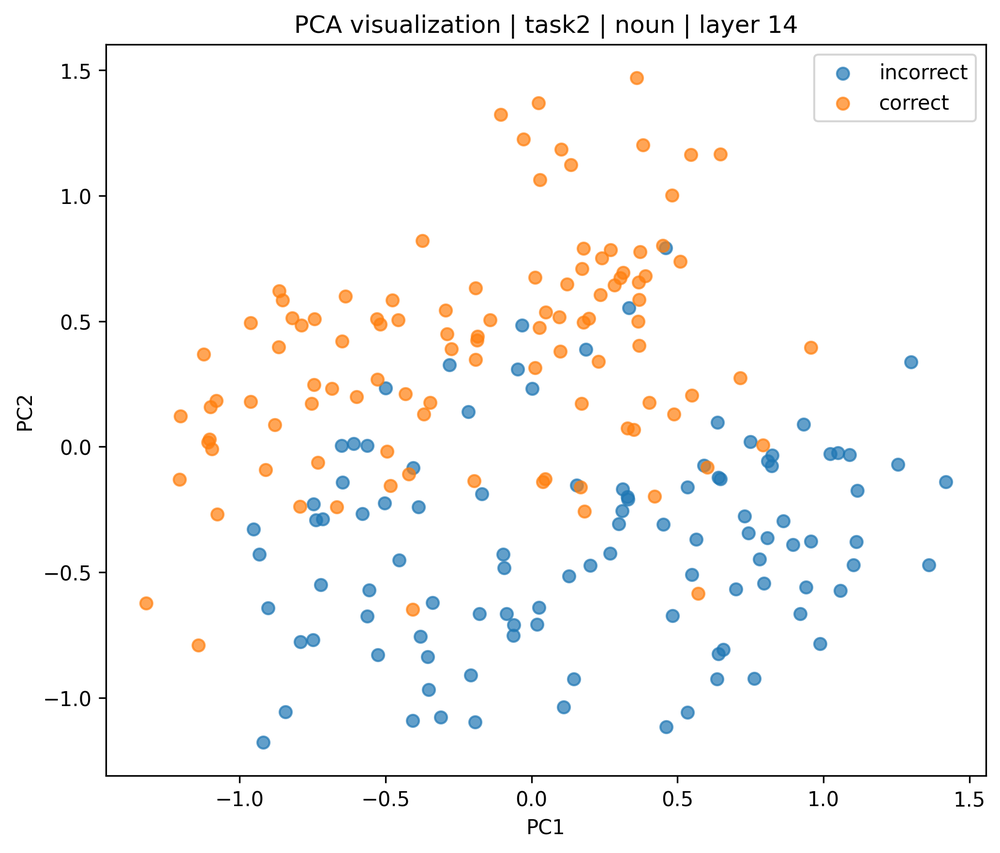}
\caption{Mistral 7B}
\end{subfigure}
\hfill
\begin{subfigure}{0.24\textwidth}
\includegraphics[width=\linewidth]{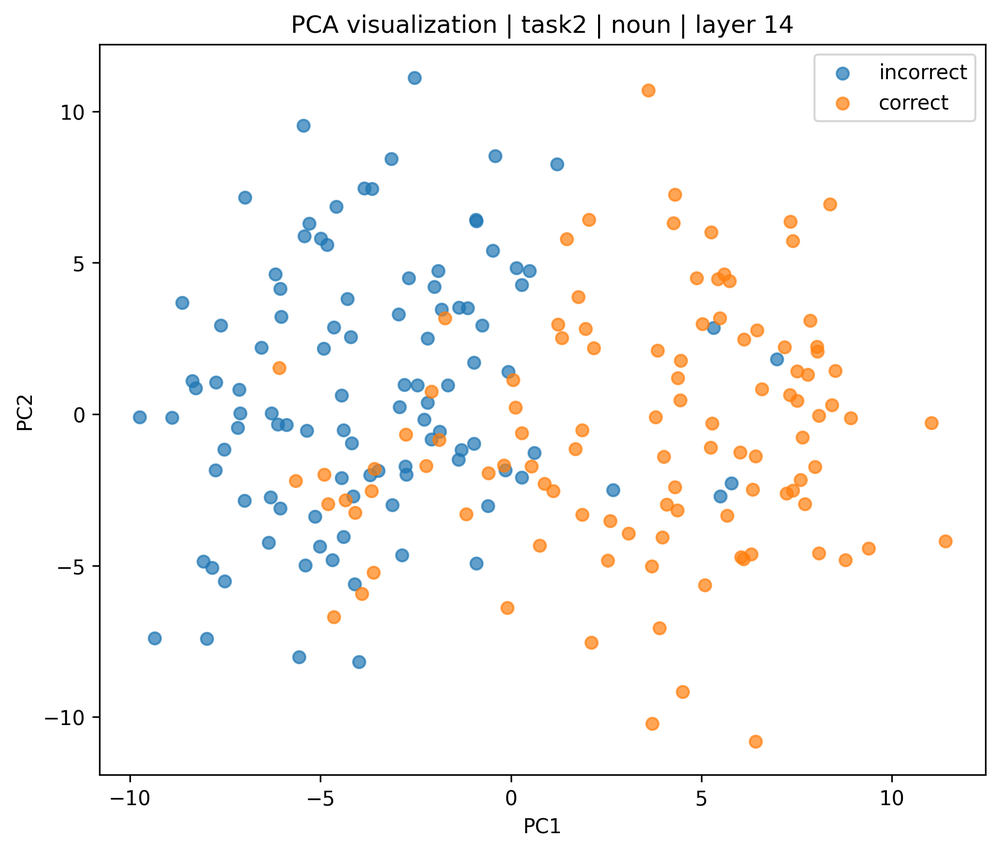}
\caption{Mixtral 8x7B}
\end{subfigure}
\hfill
\begin{subfigure}{0.24\textwidth}
\includegraphics[width=\linewidth]{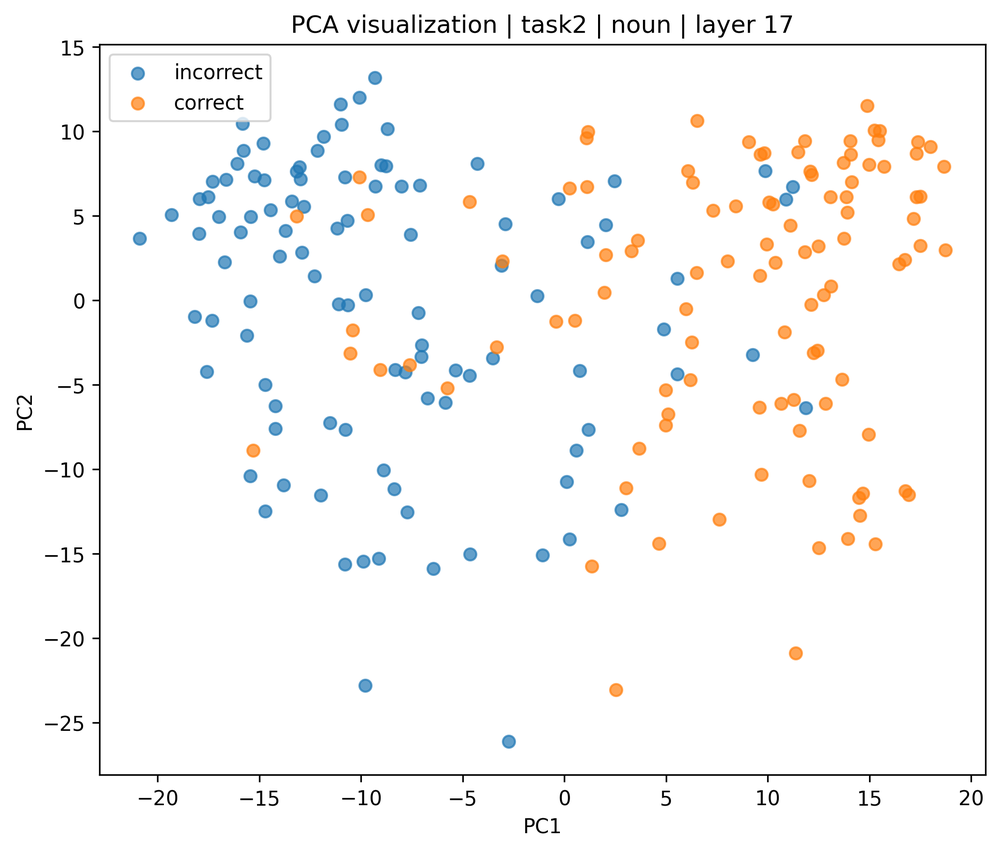}
\caption{Qwen2.5 7B}
\end{subfigure}
\hfill
\begin{subfigure}{0.24\textwidth}
\includegraphics[width=\linewidth]{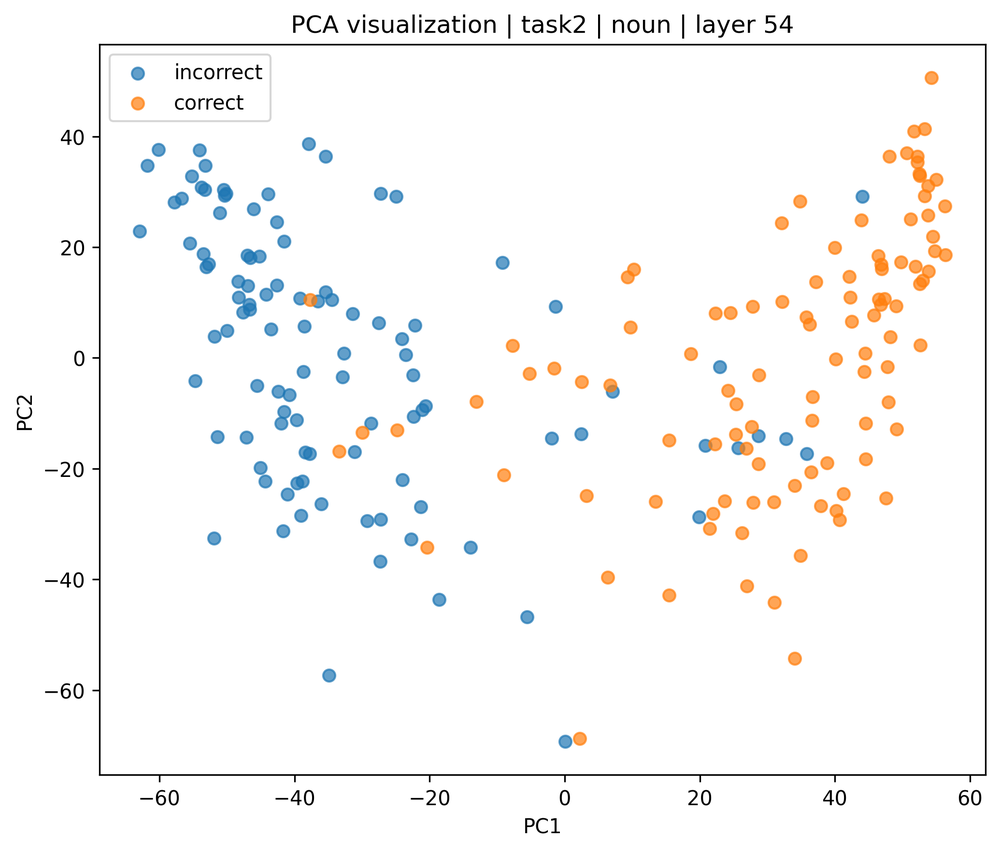}
\caption{Qwen2.5 72B}
\end{subfigure}

\caption{Interpretation task: PCA visualizations for noun representations.}
\label{fig:pca-task2-noun}
\end{figure*}

\twocolumn

\onecolumn
\section{Domain-specific DAS training at full-stop token position}

\begin{table}[H]
\centering
\caption{Domain-specific DAS IIA (mean $\pm$ std) across models. Best per domain in bold.}
\label{tab:domain-das-iia}
\small
\setlength{\tabcolsep}{4pt}
\resizebox{\textwidth}{!}{%
\begin{tabular}{lccccccc}
\toprule
Domain & LLaMA-8B & LLaMA-70B & Gemma-9B & Mistral-7B & Mixtral-8x7B & Qwen-7B & Qwen-72B \\
\midrule
Clarity     & $0.316 \pm 0.070$ & $0.351 \pm 0.022$ & $0.441 \pm 0.010$ & $0.245 \pm 0.013$ & $0.325 \pm 0.026$ & $\mathbf{0.521 \pm 0.006}$ & $0.313 \pm 0.040$ \\
Depth       & $0.343 \pm 0.010$ & $0.212 \pm 0.095$ & $0.227 \pm 0.019$ & $\mathbf{0.447 \pm 0.008}$ & $0.406 \pm 0.005$ & $0.287 \pm 0.061$ & $0.079 \pm 0.016$ \\
Light       & $0.322 \pm 0.122$ & $0.059 \pm 0.006$ & $0.524 \pm 0.015$ & $0.089 \pm 0.008$ & $0.158 \pm 0.067$ & $0.461 \pm 0.017$ & $\mathbf{0.608 \pm 0.021}$ \\
Strength    & $0.606 \pm 0.061$ & $0.508 \pm 0.043$ & $0.776 \pm 0.011$ & $0.417 \pm 0.013$ & $0.467 \pm 0.033$ & $\mathbf{0.854 \pm 0.009}$ & $0.594 \pm 0.087$ \\
Substance   & $0.309 \pm 0.059$ & $0.130 \pm 0.037$ & $0.500 \pm 0.019$ & $0.424 \pm 0.013$ & $0.319 \pm 0.036$ & $\mathbf{0.514 \pm 0.000}$ & $0.062 \pm 0.007$ \\
Taste       & $0.645 \pm 0.054$ & $0.248 \pm 0.066$ & $0.587 \pm 0.043$ & $0.662 \pm 0.037$ & $0.494 \pm 0.008$ & $0.796 \pm 0.053$ & $\mathbf{0.826 \pm 0.009}$ \\
Temperature & $0.470 \pm 0.051$ & $0.077 \pm 0.027$ & $0.583 \pm 0.021$ & $0.542 \pm 0.017$ & $0.405 \pm 0.014$ & $0.672 \pm 0.019$ & $\mathbf{0.838 \pm 0.054}$ \\
Texture     & $0.343 \pm 0.022$ & $0.219 \pm 0.084$ & $0.506 \pm 0.050$ & $\mathbf{0.810 \pm 0.035}$ & $0.613 \pm 0.062$ & $0.698 \pm 0.014$ & $0.533 \pm 0.032$ \\
\bottomrule
\end{tabular}
}
\end{table}

\section{Transfer: validation IIA using global DAS instead of domain-specific DAS}

\begin{table*}[!htbp]
\centering
\caption{Transfer IIA: global DAS evaluated on held-out domain (mean [95\% CI]). Best per domain in bold.}
\label{tab:transfer-iia}
\small
\setlength{\tabcolsep}{3pt}
\resizebox{\textwidth}{!}{%
\begin{tabular}{lccccccc}
\toprule
Domain & LLaMA-8B & LLaMA-70B & Gemma-9B & Mistral-7B & Mixtral-8x7B & Qwen-7B & Qwen-72B \\
\midrule
Clarity     & $0.500\ [0.404, 0.596]$ & $0.500\ [0.357, 0.625]$ & $0.581\ [0.484, 0.661]$ & $0.409\ [0.352, 0.466]$ & $\mathbf{0.738\ [0.680, 0.801]}$ & $0.618\ [0.542, 0.701]$ & $0.739\ [0.681, 0.792]$ \\
Depth       & $\mathbf{0.910\ [0.868, 0.952]}$ & $0.727\ [0.688, 0.767]$ & $0.924\ [0.896, 0.952]$ & $0.849\ [0.808, 0.887]$ & $0.826\ [0.781, 0.868]$ & $0.852\ [0.812, 0.893]$ & $0.721\ [0.677, 0.767]$ \\
Light       & $0.727\ [0.674, 0.776]$ & $0.240\ [0.192, 0.286]$ & $0.580\ [0.523, 0.639]$ & $0.455\ [0.400, 0.510]$ & $\mathbf{0.821\ [0.776, 0.864]}$ & $0.627\ [0.571, 0.679]$ & $0.763\ [0.717, 0.809]$ \\
Strength    & $\mathbf{0.967\ [0.953, 0.982]}$ & $0.956\ [0.938, 0.972]$ & $0.919\ [0.895, 0.942]$ & $0.595\ [0.545, 0.642]$ & $0.777\ [0.734, 0.818]$ & $0.933\ [0.908, 0.958]$ & $0.959\ [0.940, 0.975]$ \\
Substance   & $\mathbf{0.814\ [0.714, 0.900]}$ & $0.469\ [0.344, 0.594]$ & $0.643\ [0.500, 0.786]$ & $0.600\ [0.486, 0.700]$ & $0.586\ [0.471, 0.700]$ & $0.757\ [0.657, 0.857]$ & $0.771\ [0.671, 0.857]$ \\
Taste       & $0.842\ [0.763, 0.921]$ & $0.656\ [0.544, 0.744]$ & $0.830\ [0.750, 0.909]$ & $0.714\ [0.600, 0.815]$ & $0.517\ [0.400, 0.633]$ & $0.871\ [0.774, 0.952]$ & $\mathbf{0.977\ [0.942, 1.000]}$ \\
Temperature & $0.568\ [0.500, 0.627]$ & $0.235\ [0.183, 0.287]$ & $0.650\ [0.585, 0.705]$ & $0.673\ [0.601, 0.744]$ & $0.601\ [0.538, 0.664]$ & $0.735\ [0.676, 0.790]$ & $\mathbf{0.900\ [0.863, 0.933]}$ \\
Texture     & $0.713\ [0.657, 0.765]$ & $0.488\ [0.417, 0.554]$ & $0.767\ [0.714, 0.813]$ & $0.907\ [0.872, 0.943]$ & $0.809\ [0.756, 0.858]$ & $0.746\ [0.690, 0.798]$ & $\mathbf{0.924\ [0.890, 0.955]}$ \\
\bottomrule
\end{tabular}%
}
\end{table*}

\newpage
\section{Steering incorrect predictions}
\label{sec:das_steering_app}

\begin{table}[!htbp]
\centering
\small
\caption{Flip rates on initially-wrong items: mass-mean steering vs.\ DAS intervention (mean [95\% CI]). Best per row in bold.}
\label{tab:flip-rates}
\begin{tabular}{lccc}
\toprule
Model & Wrong/Total & Mass-Mean flip & DAS flip \\
\midrule
Gemma-9B     & 1140/8591 & $0.750\ [0.660, 0.830]$ & $\mathbf{0.860\ [0.790, 0.920]}$ \\
LLaMA-8B     & 1734/8591 & $0.540\ [0.450, 0.640]$ & $\mathbf{0.710\ [0.620, 0.790]}$ \\
Qwen-7B      & 1808/8591 & $\mathbf{0.620\ [0.530, 0.710]}$ & $0.610\ [0.520, 0.700]$ \\
Mistral-7B   & 2079/8591 & $\mathbf{0.640\ [0.540, 0.730]}$ & $0.450\ [0.350, 0.550]$ \\
Mixtral-8x7B & 1798/8591 & $\mathbf{0.720\ [0.630, 0.810]}$ & $0.610\ [0.510, 0.700]$ \\
LLaMA-70B    & 1222/8591 & $0.570\ [0.470, 0.670]$ & $\mathbf{0.650\ [0.560, 0.740]}$ \\
Qwen-72B     & 1043/8591 & $0.880\ [0.810, 0.940]$ & $\mathbf{0.910\ [0.850, 0.960]}$ \\
\bottomrule
\end{tabular}
\end{table}

\section{LLM as a judge}
\label{sec:generation_steering}

\begin{table}[H]
\centering
\caption{Literal $\rightarrow$ Metaphor generation flip rate by steering scale $\alpha$ and method (MM = mass-mean, DAS). Mean [95\% CI]. Degenerate outputs counted as failed flips, few-shot prompts excluded. Only pairs where the judge gets the baseline explanation correct are considered (e.g. the meaning of ``warm'' in ``the coffee is warm'' is a model explanation such that the judge classifies it as ``literal'').}
\label{tab:literal-to-metaphor-ci}
\scriptsize
\setlength{\tabcolsep}{3pt}
\begin{tabular}{cl ccccccc}
\toprule
$\alpha$ & Method & LLaMA-8B & LLaMA-70B & Gemma-9B & Mistral-7B & Mixtral-8x7B & Qwen-7B & Qwen-72B \\
\midrule
\multirow{2}{*}{1}  & MM  & $0.05\ [0.00, 0.10]$ & $0.03\ [0.00, 0.08]$ & $0.02\ [0.00, 0.05]$ & $0.03\ [0.00, 0.08]$ & $0.05\ [0.00, 0.11]$ & $0.03\ [0.00, 0.09]$ & $0.08\ [0.02, 0.15]$ \\
                    & DAS & $0.02\ [0.00, 0.05]$ & $0.03\ [0.00, 0.08]$ & $0.00\ [0.00, 0.00]$ & $0.05\ [0.00, 0.11]$ & $0.02\ [0.00, 0.05]$ & $0.03\ [0.00, 0.09]$ & $0.02\ [0.00, 0.05]$ \\
\midrule
\multirow{2}{*}{2}  & MM  & $0.12\ [0.05, 0.20]$ & $0.05\ [0.00, 0.11]$ & $0.11\ [0.05, 0.18]$ & $0.11\ [0.03, 0.19]$ & $0.11\ [0.03, 0.19]$ & $0.12\ [0.05, 0.22]$ & $0.09\ [0.03, 0.17]$ \\
                    & DAS & $0.03\ [0.00, 0.08]$ & $0.08\ [0.02, 0.16]$ & $0.03\ [0.00, 0.08]$ & $0.20\ [0.11, 0.31]$ & $0.03\ [0.00, 0.08]$ & $0.09\ [0.02, 0.16]$ & $0.05\ [0.00, 0.11]$ \\
\midrule
\multirow{2}{*}{5}  & MM  & $0.44\ [0.32, 0.56]$ & $0.34\ [0.23, 0.45]$ & $0.33\ [0.23, 0.44]$ & $0.52\ [0.39, 0.64]$ & $0.38\ [0.27, 0.50]$ & $0.26\ [0.16, 0.38]$ & $0.29\ [0.20, 0.40]$ \\
                    & DAS & $0.34\ [0.22, 0.46]$ & $0.28\ [0.17, 0.41]$ & $0.12\ [0.06, 0.21]$ & $0.19\ [0.09, 0.28]$ & $0.16\ [0.08, 0.25]$ & $0.47\ [0.34, 0.59]$ & $0.15\ [0.08, 0.25]$ \\
\midrule
\multirow{2}{*}{10} & MM  & $0.20\ [0.12, 0.31]$ & $0.47\ [0.36, 0.59]$ & $0.55\ [0.42, 0.67]$ & $0.34\ [0.23, 0.45]$ & $0.47\ [0.34, 0.59]$ & $0.57\ [0.45, 0.69]$ & $0.11\ [0.05, 0.18]$ \\
                    & DAS & $0.25\ [0.15, 0.37]$ & $0.00\ [0.00, 0.00]$ & $0.03\ [0.00, 0.08]$ & $0.02\ [0.00, 0.05]$ & $0.06\ [0.02, 0.12]$ & $0.40\ [0.28, 0.52]$ & $0.43\ [0.31, 0.55]$ \\
\bottomrule
\end{tabular}
\end{table}

\begin{table}[H]
\centering
\caption{Metaphor $\rightarrow$ Literal generation flip rate by steering scale $\alpha$ and method (MM = mass-mean, DAS). Mean [95\% CI]. Degenerate outputs counted as failed flips; few-shot prompts excluded, only baseline-correct items.}
\label{tab:metaphor-to-literal-ci}
\scriptsize
\setlength{\tabcolsep}{3pt}
\begin{tabular}{cl ccccccc}
\toprule
$\alpha$ & Method & LLaMA-8B & LLaMA-70B & Gemma-9B & Mistral-7B & Mixtral-8x7B & Qwen-7B & Qwen-72B \\
\midrule
\multirow{2}{*}{1}  & MM  & $0.03\ [0.00, 0.07]$ & $0.01\ [0.00, 0.04]$ & $0.05\ [0.01, 0.11]$ & $0.06\ [0.01, 0.11]$ & $0.01\ [0.00, 0.04]$ & $0.04\ [0.00, 0.09]$ & $0.00\ [0.00, 0.00]$ \\
                    & DAS & $0.00\ [0.00, 0.00]$ & $0.01\ [0.00, 0.04]$ & $0.05\ [0.01, 0.11]$ & $0.00\ [0.00, 0.00]$ & $0.00\ [0.00, 0.00]$ & $0.04\ [0.00, 0.09]$ & $0.00\ [0.00, 0.00]$ \\
\midrule
\multirow{2}{*}{2}  & MM  & $0.04\ [0.00, 0.09]$ & $0.04\ [0.00, 0.09]$ & $0.11\ [0.04, 0.18]$ & $0.04\ [0.00, 0.10]$ & $0.07\ [0.01, 0.14]$ & $0.08\ [0.03, 0.14]$ & $0.00\ [0.00, 0.00]$ \\
                    & DAS & $0.01\ [0.00, 0.04]$ & $0.05\ [0.01, 0.11]$ & $0.08\ [0.03, 0.14]$ & $0.00\ [0.00, 0.00]$ & $0.01\ [0.00, 0.04]$ & $0.07\ [0.01, 0.13]$ & $0.03\ [0.00, 0.07]$ \\
\midrule
\multirow{2}{*}{5}  & MM  & $0.03\ [0.00, 0.07]$ & $0.15\ [0.07, 0.23]$ & $0.25\ [0.16, 0.36]$ & $0.28\ [0.18, 0.39]$ & $0.24\ [0.14, 0.34]$ & $0.21\ [0.12, 0.30]$ & $0.08\ [0.03, 0.15]$ \\
                    & DAS & $0.09\ [0.04, 0.16]$ & $0.16\ [0.08, 0.25]$ & $0.12\ [0.05, 0.20]$ & $0.10\ [0.04, 0.17]$ & $0.11\ [0.04, 0.19]$ & $0.12\ [0.05, 0.18]$ & $0.19\ [0.11, 0.27]$ \\
\midrule
\multirow{2}{*}{10} & MM  & $0.03\ [0.00, 0.07]$ & $0.00\ [0.00, 0.00]$ & $0.13\ [0.07, 0.21]$ & $0.35\ [0.24, 0.46]$ & $0.33\ [0.21, 0.44]$ & $0.01\ [0.00, 0.04]$ & $0.03\ [0.00, 0.07]$ \\
                    & DAS & $0.00\ [0.00, 0.00]$ & $0.05\ [0.01, 0.11]$ & $0.03\ [0.00, 0.07]$ & $0.06\ [0.01, 0.11]$ & $0.24\ [0.14, 0.34]$ & $0.05\ [0.01, 0.11]$ & $0.25\ [0.16, 0.35]$ \\
\bottomrule
\end{tabular}
\end{table}

\section{Example generations}

\begin{table}[H]
\centering
\small
\begin{tabular}{@{}lp{0.62\linewidth}c@{}}
\toprule
 & \textbf{Output} & \textbf{Steered?} \\
\midrule
\multicolumn{3}{@{}l}{\textbf{Metaphorical $\rightarrow$ Literal}} \\
\midrule
\multicolumn{3}{@{}l}{\textbf{clear thinker} (CLARITY)} \\
Baseline & easy to understand and free from confusion. & --- \\
MM       & easy to understand and free from confusion. & \ding{55} \\
DAS      & easy to understand and not ambiguous. & \ding{55} \\
\midrule
\multicolumn{3}{@{}l}{\textbf{smooth player} (TEXTURE)} \\
Baseline & having a calm and confident manner. & --- \\
MM       & having a calm and peaceful appearance. & \ding{55} \\
DAS      & having a flat, even surface, and not necessarily emotionally complex. & \ding{51} \\
\midrule
\multicolumn{3}{@{}l}{\textbf{Literal $\rightarrow$ Metaphorical}} \\
\midrule
\multicolumn{3}{@{}l}{\textbf{rough landscape} (TEXTURE)} \\
Baseline & having a harsh, uneven surface. & --- \\
MM       & having an uneven surface with many obstacles. & \ding{55} \\
DAS      & having an uneven surface with many obstacles. & \ding{55} \\
\midrule
\multicolumn{3}{@{}l}{\textbf{clear broth} (CLARITY)} \\
Baseline & transparent and free from impurities. & --- \\
MM       & transparent and easy to understand. & \ding{51} \\
DAS      & transparent and free from impurities. & \ding{55} \\
\bottomrule
\end{tabular}
\caption{Qualitative examples of steering outcomes for mass-mean (MM) and DAS interventions. \ding{51} indicates the judge's label flipped to the steered direction. \ding{55} indicates no flip.}
\label{tab:steering-examples}
\end{table}

\section{Matched-$n$ control}

\begin{table}[H]
  \centering
  \begin{tabular}{lcccc}
  \hline
  Concept & Full $n$ & Matched $n$ & Domain-specific & $M_n - \text{Dom}$ [95\% CI] \\
   & (global DAS) & ($M_n$, global DAS) & DAS (Dom) & \\
  \hline
  clarity      & 0.581 & 0.513 & 0.441 & $+0.073$ $[+0.003, +0.145]$ \\
  depth        & 0.924 & 0.855 & 0.227 & $+0.628$ $[+0.598, +0.662]$ \\
  light        & 0.580 & 0.412 & 0.524 & $-0.113$ $[-0.158, -0.068]$ \\
  strength     & 0.919 & 0.850 & 0.776 & $+0.074$ $[+0.046, +0.100]$ \\
  substance    & 0.643 & 0.524 & 0.500 & $+0.024$ $[-0.095, +0.143]$ \\
  taste        & 0.830 & 0.750 & 0.587 & $+0.163$ $[+0.083, +0.239]$ \\
  temperature  & 0.650 & 0.447 & 0.583 & $-0.135$ $[-0.187, -0.084]$ \\
  texture      & 0.767 & 0.587 & 0.506 & $+0.080$ $[+0.030, +0.132]$ \\
  \hline
  \end{tabular}
  \caption{We retrain the global DAS matrix on a number of samples for each domain, matching the smallest domain-specific training domain ($n=255$). For 6/8 cells, we achieve a comparable transfer IIA score to the global DAS matrix where no such filter is applied. Results reported for Gemma 2 9B.}
  \label{tab:matched-n}
  \end{table}

\twocolumn
\section{Steering meaning}
\label{sec:NIEs}

\subsection{NIE calculation}
For \textit{identification} task, we first compute $PD^+$ and $PD^-$ as the average probability differences $P(\text{metaphoric}) - P(\text{literal})$ over examples from the metaphoric($+$) and literal($-$) classes, respectively. After intervention, we obtain $PD^{+}_*$ and $PD^{-}_*$ in the same way.
The NIE is then computed as:
\[
\text{NIE}_{\text{lit}\rightarrow\text{met}} = 
\frac{PD^{-}_* - PD^-}{PD^+ - PD^-}\]

\[\text{NIE}_{\text{met}\rightarrow\text{lit}} = 
\frac{PD^{+}_* - PD^+}{PD^- - PD^+}.
\]

Therefore, an NIE of 0 indicates no causal effect, while an NIE of 1 corresponds to a complete shift in model predictions.

\clearpage
\onecolumn
\subsection{Full NIE results}
\begin{table}[H]
\centering
\scriptsize
\setlength{\tabcolsep}{3.5pt}
\renewcommand{\arraystretch}{1.12}

\resizebox{\textwidth}{!}{
\begin{tabular}{ll||cc|cc|cc|cc}
\toprule
\multirow{2}{*}{train/test} &
\multirow{2}{*}{probe}
& \multicolumn{2}{c|}{LLaMA-8B}
& \multicolumn{2}{c|}{LLaMA-70B}
& \multicolumn{2}{c|}{Gemma-9B}
& \multicolumn{2}{c}{Gemma-27B} \\
\cmidrule(lr){3-4}\cmidrule(lr){5-6}\cmidrule(lr){7-8}\cmidrule(lr){9-10}
& &
inapt$\rightarrow$apt & apt$\rightarrow$inapt
& inapt$\rightarrow$apt & apt$\rightarrow$inapt
& inapt$\rightarrow$apt & apt$\rightarrow$inapt
& inapt$\rightarrow$apt & apt$\rightarrow$inapt \\
\midrule

\multirow{2}{*}{Adj/Adj}
& LogReg  & .038 & .051 & .033 & .028 & \textbf{.176} & \textbf{.211} & \textbf{.293} & \textbf{.774} \\
& MM-diff & \textbf{.227} & \textbf{.471} & \textbf{.183} & \textbf{.168} & .149 & .097 & .262 & .387 \\
\midrule

\multirow{2}{*}{Noun/Noun}
& LogReg  & .025 & .040 & .029 & .029 & \textbf{.220} & .060 & \textbf{.227} & .282 \\
& MM-diff & \textbf{.195} & \textbf{.355} & \textbf{.216} & \textbf{.126} & .212 & \textbf{.110} & .045 & \textbf{.304} \\
\midrule

\multirow{2}{*}{Adj/Noun}
& LogReg  & .026 & .032 & .029 & .031 & \textbf{.211} & \textbf{.297} & \textbf{.076} & \textbf{.763} \\
& MM-diff & \textbf{.138} & \textbf{.254} & \textbf{.189} & \textbf{.124} & .177 & .223 & .050 & .194 \\
\midrule

\multirow{2}{*}{Noun/Adj}
& LogReg  & .028 & .040 & .019 & .015 & .076 & .072 & \textbf{.449} & .429 \\
& MM-diff & \textbf{.230} & \textbf{.666} & \textbf{.121} & \textbf{.171} & \textbf{.110} & \textbf{.131} & .230 & \textbf{.497} \\

\midrule\midrule

\multirow{2}{*}{train/test} &
\multirow{2}{*}{probe}
& \multicolumn{2}{c|}{Mistral-7B}
& \multicolumn{2}{c|}{Mixtral-8x7B}
& \multicolumn{2}{c|}{Qwen-7B}
& \multicolumn{2}{c}{Qwen-72B} \\
\cmidrule(lr){3-4}\cmidrule(lr){5-6}\cmidrule(lr){7-8}\cmidrule(lr){9-10}
& &
inapt$\rightarrow$apt & apt$\rightarrow$inapt
& inapt$\rightarrow$apt & apt$\rightarrow$inapt
& inapt$\rightarrow$apt & apt$\rightarrow$inapt
& inapt$\rightarrow$apt & apt$\rightarrow$inapt \\
\midrule

\multirow{2}{*}{Adj/Adj}
& LogReg  & .014 & .095 & .075 & .178 & .035 & .054 & .100 & .084 \\
& MM-diff & \textbf{.075} & \textbf{.449} & \textbf{.093} & \textbf{.292} & \textbf{.676} & \textbf{.578} & \textbf{.251} & \textbf{.503} \\
\midrule

\multirow{2}{*}{Noun/Noun}
& LogReg  & .007 & \textbf{.033} & .013 & .042 & .010 & .031 & .097 & \textbf{.072} \\
& MM-diff & \textbf{.023} & .025 & \textbf{.031} & \textbf{.051} & \textbf{.467} & \textbf{.081} & \textbf{.243} & .065 \\
\midrule

\multirow{2}{*}{Adj/Noun}
& LogReg  & .006 & .030 & .012 & .123 & .050 & .056 & .097 & .040 \\
& MM-diff & \textbf{.024} & \textbf{.153} & \textbf{.015} & \textbf{.159} & \textbf{.397} & \textbf{.561} & \textbf{.167} & \textbf{.397} \\
\midrule

\multirow{2}{*}{Noun/Adj}
& LogReg  & .006 & .056 & .042 & .123 & .038 & .031 & .072 & .060 \\
& MM-diff & \textbf{.054} & \textbf{.520} & \textbf{.093} & \textbf{.367} & \textbf{.553} & \textbf{.353} & \textbf{.285} & \textbf{.318} \\

\bottomrule
\end{tabular}
}

\caption{Normalized indirect effects (NIEs) of mass-mean difference (MM-diff) and logistic regression (LogReg) probes on the \textit{identification} task for eight models. Higher NIE values indicate stronger causal steering effects. Higher scores are highlighted in bold.}
\label{tab:nie_task1}
\end{table}

\begin{table}[H]
\centering
\scriptsize
\setlength{\tabcolsep}{3.5pt}
\renewcommand{\arraystretch}{1.12}

\resizebox{\textwidth}{!}{
\begin{tabular}{ll||cc|cc|cc|cc}
\toprule
\multirow{2}{*}{train/test} &
\multirow{2}{*}{probe}
& \multicolumn{2}{c|}{LLaMA-8B}
& \multicolumn{2}{c|}{LLaMA-70B}
& \multicolumn{2}{c|}{Gemma-9B}
& \multicolumn{2}{c}{Gemma-27B} \\
\cmidrule(lr){3-4}\cmidrule(lr){5-6}\cmidrule(lr){7-8}\cmidrule(lr){9-10}
& &
inapt$\rightarrow$apt & apt$\rightarrow$inapt
& inapt$\rightarrow$apt & apt$\rightarrow$inapt
& inapt$\rightarrow$apt & apt$\rightarrow$inapt
& inapt$\rightarrow$apt & apt$\rightarrow$inapt \\
\midrule

\multirow{2}{*}{Noun}
& LogReg  & .011 & .014 & .001 & .005 & .009 & .031 & .038 & .144 \\
& MM-diff & \textbf{.355} & \textbf{.564} & \textbf{.126} & \textbf{.305} & \textbf{.155} & \textbf{.353} & \textbf{.304} & \textbf{.656} \\

\midrule\midrule

\multirow{2}{*}{train/test} &
\multirow{2}{*}{probe}
& \multicolumn{2}{c|}{Mistral-7B}
& \multicolumn{2}{c|}{Mixtral-8x7B}
& \multicolumn{2}{c|}{Qwen-7B}
& \multicolumn{2}{c}{Qwen-72B} \\
\cmidrule(lr){3-4}\cmidrule(lr){5-6}\cmidrule(lr){7-8}\cmidrule(lr){9-10}
& &
inapt$\rightarrow$apt & apt$\rightarrow$inapt
& inapt$\rightarrow$apt & apt$\rightarrow$inapt
& inapt$\rightarrow$apt & apt$\rightarrow$inapt
& inapt$\rightarrow$apt & apt$\rightarrow$inapt \\
\midrule

\multirow{2}{*}{Noun}
& LogReg  & .003 & .013 & .003 & .017 & .002 & .008 & .007 & .033 \\
& MM-diff & \textbf{.025} & \textbf{.165} & \textbf{.051} & \textbf{.238} & \textbf{.081} & \textbf{.261} & \textbf{.065} & \textbf{.344} \\

\bottomrule
\end{tabular}
}
\caption{Normalized indirect effects (NIEs) of mass-mean difference (MM-diff) and logistic regression (LogReg) probes on the \textit{interpretation} task for eight models. Probes are trained and tested on our Noun dataset. Higher NIE values indicate stronger causal steering effects. Higher scores are highlighted in bold.}
\label{tab:nie_task2}
\end{table}

\subsection{Flip rate results}

\begin{table}[H]
\centering
\scriptsize
\setlength{\tabcolsep}{4pt}
\renewcommand{\arraystretch}{1.12}

\resizebox{\textwidth}{!}{
\begin{tabular}{lllclrrr}
\toprule
Task & Model & Data & Layer & Original label
& Before & After & Flip rate \\
\midrule

\multirow{12}{*}{T1}
& LLaMA-3.1-8B  & Noun & 11 & metaphor & 27 & 13 & 51.85\% \\
& LLaMA-3.1-8B  & Noun & 11 & literal  & 28 &  7 & 75.00\% \\
& LLaMA-3.1-8B  & Adj  & 11 & metaphor & 21 & 14 & 33.33\% \\
& LLaMA-3.1-8B  & Adj  & 11 & literal  &  7 &  2 & 71.43\% \\
\cmidrule(lr){2-8}
& Gemma-2-27B   & Noun & 20 & literal  & 23 & 18 & 21.74\% \\
& Gemma-2-27B   & Adj  & 20 & literal  & 17 & 12 & 29.41\% \\
\cmidrule(lr){2-8}
& Qwen2.5-72B   & Noun & 54 & literal  & 26 & 18 & 30.77\% \\
& Qwen2.5-72B   & Adj  & 54 & literal  & 18 &  8 & 55.56\% \\
\cmidrule(lr){2-8}
& Mixtral-8x7B  & Noun & 14 & literal  & 20 & 14 & 30.00\% \\
& Mixtral-8x7B  & Adj  & 14 & literal  & 13 & 10 & 23.08\% \\
\cmidrule(lr){2-8}
& LLaMA-3.1-70B & Noun & 29 & literal  & 25 & 22 &  8.00\% \\
& LLaMA-3.1-70B & Adj  & 29 & literal  & 18 & 16 & 11.11\% \\

\midrule

\multirow{4}{*}{T2}
& Gemma-2-27B  & Noun & 20 & inapt & 26 & 21 & 19.23\% \\
& Gemma-2-9B   & Noun & 20 & inapt & 26 & 23 & 11.54\% \\
& Qwen2.5-72B  & Noun & 54 & inapt & 25 & 24 &  4.00\% \\
& Mixtral-8x7B & Noun & 14 & inapt & 27 & 25 &  7.41\% \\

\bottomrule
\end{tabular}
}

\caption{Steering results using mass-mean probes on the identification
(T1) and interpretation (T2) tasks. We intervene on correctly classified
test examples and steer them toward the opposite label. ``Before'' and
``After'' indicate the numbers of examples retaining the original label
before and after steering, respectively. Flip rate denotes the proportion
of originally correct predictions that are flipped to the opposite label.}

\label{tab:mm_flip}
\end{table}
\twocolumn

\subsection{Prompt format}

All models use the same plain-continuation meaning-format prompt with five hardcoded
few-shot examples demonstrating free-form meaning descriptions (no classification
labels). The test query asks for the meaning of the adjective in context:
\\
\
\\
The welcome was warm. The meaning of 'warm' is: affectionate and emotionally inviting.\\
The lamp was bright. The meaning of 'bright' is: emitting a strong, clear light.\\
The novel was deep. The meaning of 'deep' is: intellectually complex and rich in meaning.\\
The stone was smooth. The meaning of 'smooth' is: having an even surface with no roughness.\\
The defeat was bitter. The meaning of 'bitter' is: painful, hard to accept, and leaving resentment.\\
The [noun] was [adj]. The meaning of '[adj]' is:

No chat template is applied to any model. The hook patches the hidden state at the
last token position at every forward step during generation. The token position is the same token position used for DAS training.

\section{Prompt for LLM judge}
\label{sec:judge-prompt}

You are classifying whether a meaning description conveys a literal or metaphorical sense of an adjective.\\
\\
\
A LITERAL meaning describes a physical, concrete, or sensory property (e.g. temperature, texture, weight, brightness).\\
A METAPHORICAL meaning describes an abstract, figurative, or emotional property (e.g. mood, personality, intellectual quality).
\\
\
\\
Reply with exactly one word: either "literal" or "metaphorical".\\
No explanation. Do not repeat the sentence.\\
Sentence: "The {noun} was {adj}."\\
Adjective: '{adj}'  \\
Meaning description: "{first sentence of completion}"\\
Is this meaning literal or metaphorical?

\section{Dataset Usage}
We use datasets from \citet{liu-etal-2022-testing, gutierrez-etal-2016-literal, tsvetkov2014metaphor}. While we slight alter the format of these datasets, all are used for their original purpose, namely metaphor research. All datasets are freely available in the public domain.

\section{Computational resources}
All experiments took a total of about 125 hours on an H100 GPU. Hyperparameters checked include steering strength $\alpha \in \{1,2,5,10\}$ for generation experiments and subspace dimension $k \in \{4, 8, 16, 32\}$ for DAS matrix training. 

\section{Use of AI assistants}
AI assistants are used for code debugging, generating the prompt for the generation experiment, and generating tables. Numbers are double-checked manually.

\end{document}